\documentclass[10pt,twocolumn,letterpaper]{article}

\usepackage{iccv}
\usepackage{times}
\usepackage{epsfig}
\usepackage{graphicx}
\usepackage{amsmath}
\usepackage{amssymb}

\usepackage{import}
\usepackage{bm}
\usepackage{multirow}
\usepackage{caption}

\usepackage{authblk}

\newcommand{\RR}{\mathbb{R}}

\newcommand{\ba}{\mathbf{a}}

\newcommand{\bd}{\mathbf{d}}

\newcommand{\bl}{\mathbf{l}}
\newcommand{\bp}{\mathbf{p}}

\newcommand{\bn}{\mathbf{n}}

\newcommand{\bt}{\mathbf{t}}

\newcommand{\bx}{\mathbf{x}}

\newcommand{\bC}{\mathbf{C}}

\newcommand{\bR}{\mathbf{R}}

\usepackage[pagebackref=true,breaklinks=true,letterpaper=true,colorlinks,bookmarks=false]{hyperref}

\iccvfinalcopy 

\def\iccvPaperID{614} 

\ificcvfinal\fi
\begin{document}

\title{Dense Multi-view 3D-reconstruction Without Dense Correspondences}

\author[1]{Yvain \textsc{Qu{\'e}au} }
\author[2,3]{Jean \textsc{M{\'e}lou}}
\author[2]{Jean-Denis \textsc{Durou}}
\author[1]{Daniel \textsc{Cremers}}

\affil[1]{Technical University of Munich, Germany}
\affil[2]{University of Toulouse, France}
\affil[3]{Mikros Image, Levallois-Perret, France}


\twocolumn[{%
\renewcommand\twocolumn[1][]{#1}%
\maketitle
\begin{center}
    \setlength{\tabcolsep}{0.1em} 
    {\renewcommand{\arraystretch}{0.6}
      \begin{tabular}{cc}
        \includegraphics[height = 0.25\linewidth]{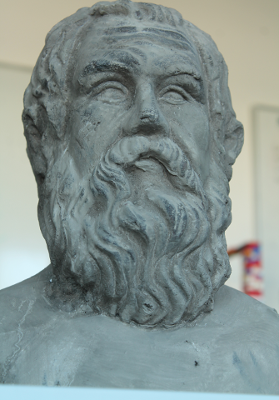} \hspace*{-2.75em}\includegraphics[height = 0.05\linewidth]{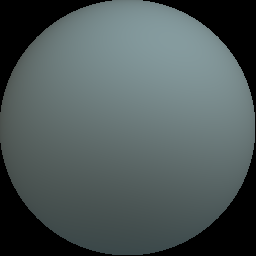}\quad
        \includegraphics[height = 0.25\linewidth]{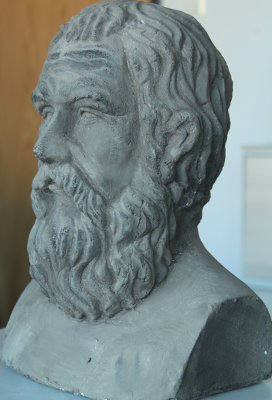} \hspace*{-2.75em}\includegraphics[height = 0.05\linewidth]{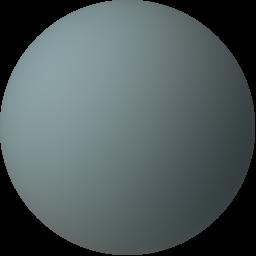} \qquad&\qquad 
        \includegraphics[height = 0.25\linewidth]{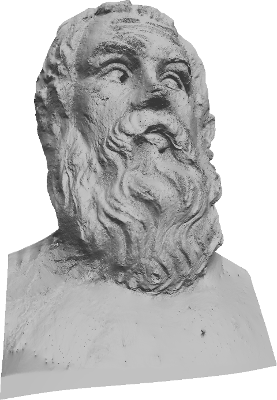}\quad
        \includegraphics[height = 0.25\linewidth]{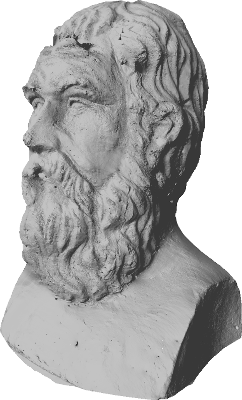} \\
        {\small $N$ multi-view input real images with illumination} \qquad&\qquad {\small Shape-from-shading without any regularization nor initial estimate}                
      \end{tabular}
    }
    \captionof{figure}{We show how to solve shape-from-shading under natural illumination without regularization, to capture the finest level of detail. When sparse correspondences across multi-view images are available (here, we used $N=4$ images from the ``Sokrates'' dataset~\cite{Zollhoefer2015}), unambiguous 3D-reconstruction is achieved. The difficulty of dense matching is thus circumvented.}
    \label{fig:teaser}
\end{center}%
}]

\begin{abstract}
\vspace*{-1em}
  We introduce a variational method for multi-view shape-from-shading
  under natural illumination. The key idea is to couple PDE-based
  solutions for single-image based shape-from-shading problems across
  multiple images and multiple color channels by means of a
  variational formulation.  Rather than alternatingly solving the
  individual SFS problems and optimizing the consistency across images
  and channels which is known to lead to suboptimal results, we
  propose an efficient solution of the coupled problem by means of an
  ADMM algorithm.  In numerous experiments on both simulated and real
  imagery, we demonstrate that the proposed fusion of multiple-view
  reconstruction and shape-from-shading provides highly accurate dense
  reconstructions without the need to compute dense correspondences.
  With the proposed variational integration across multiple views 
  shape-from-shading techniques become applicable to challenging
  real-world reconstruction problems, giving rise to highly detailed
  geometry even in areas of smooth brightness variation and lacking
  texture.  
\end{abstract}


\section{Introduction}

\subsection{Multi-view Shape-from-shading}

Over the decades the reconstruction of dense 3D geometry from images
has been tackled in numerous ways.  Two of the most popular strategies
are the reconstruction from multiple views using the notion of color
or feature correspondence and the reconstruction of shaded objects
using the technique of shape-from-shading.  Both approaches are in
many ways complementary, both have their strengths and limitations.
While the fusion of these complementary concepts in a single
reconstruction algorithm bears great promise, to date this challenge
has remained unsolved and  convicing experimental realizations have
remained elusive.  In this work, we will review existing efforts and
propose a novel solution to this challenge.

\subsection{Related Work}

\paragraph{Multi-view stereo reconstruction. }

Multi-view stereo reconstruction (MVS)~\cite{Furukawa2015} is among
the most powerful techniques to recover 3D geometry from multiple
real-world images.  The key idea is to exploit the fact that 3D points
are likely to be on the (Lambertian) object surface if the projection
into various cameras gives rise to a consistent color, patch or
feature value.  The arising photo-consistency-weighted minimal surface
problems can be optimized using techniques such as graph cuts
\cite{Vogiatzis2005} or convex relaxation
\cite{Kolev2009}.  Despite its enormous popularity for
real-world reconstruction, multi-view stereo methods have several
well-known shortcomings. Firstly, the estimation of dense
correspondences is computationally challenging~\cite{Tola2010}.
Secondly, in the absense of color variations (textureless areas), the
color consistency assumption degenerates leading to a need for
regularity or smoothness assumptions -- the resulting
photoconsistency-weighted minimal surface formulations degenerate to
Euclidean minimal surface problems which exhibit a shrinking bias that
leads to the loss of concavities, indentations and other fine-scale
geometric details.

\paragraph{Shape-from-shading. }

\begin{figure}[!ht]
\centering
\def\svgwidth{0.65\linewidth}
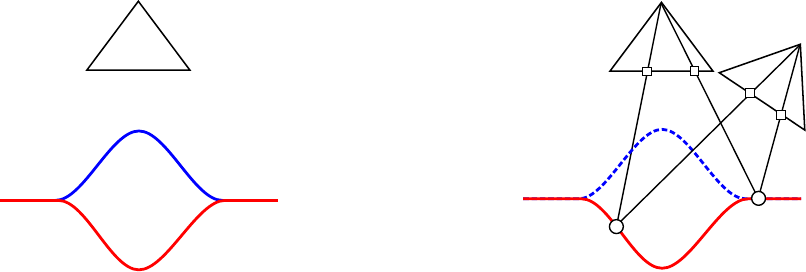 
\caption{Shape-from-shading suffers from the concave$/$convex ambiguity (left). We introduce a practical approach to SFS under natural illumination, which achieves unambiguous 3D-reconstruction when sparse correspondences between multi-view images are available (right).}
\label{fig:sketch}
\end{figure}

\begin{figure*}[!ht]
\centering
\begin{tabular}{cccc}
\includegraphics[height=0.17\linewidth]{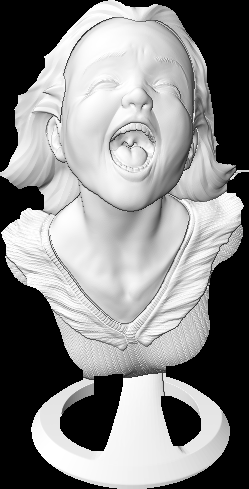}
\hspace*{-0.0cm}\includegraphics[height=0.05\linewidth]{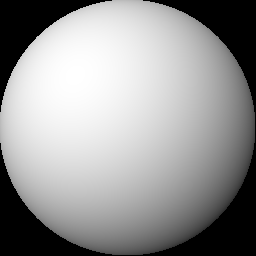} &
\includegraphics[height=0.17\linewidth,trim = 10em 6em 10em 6em,clip]{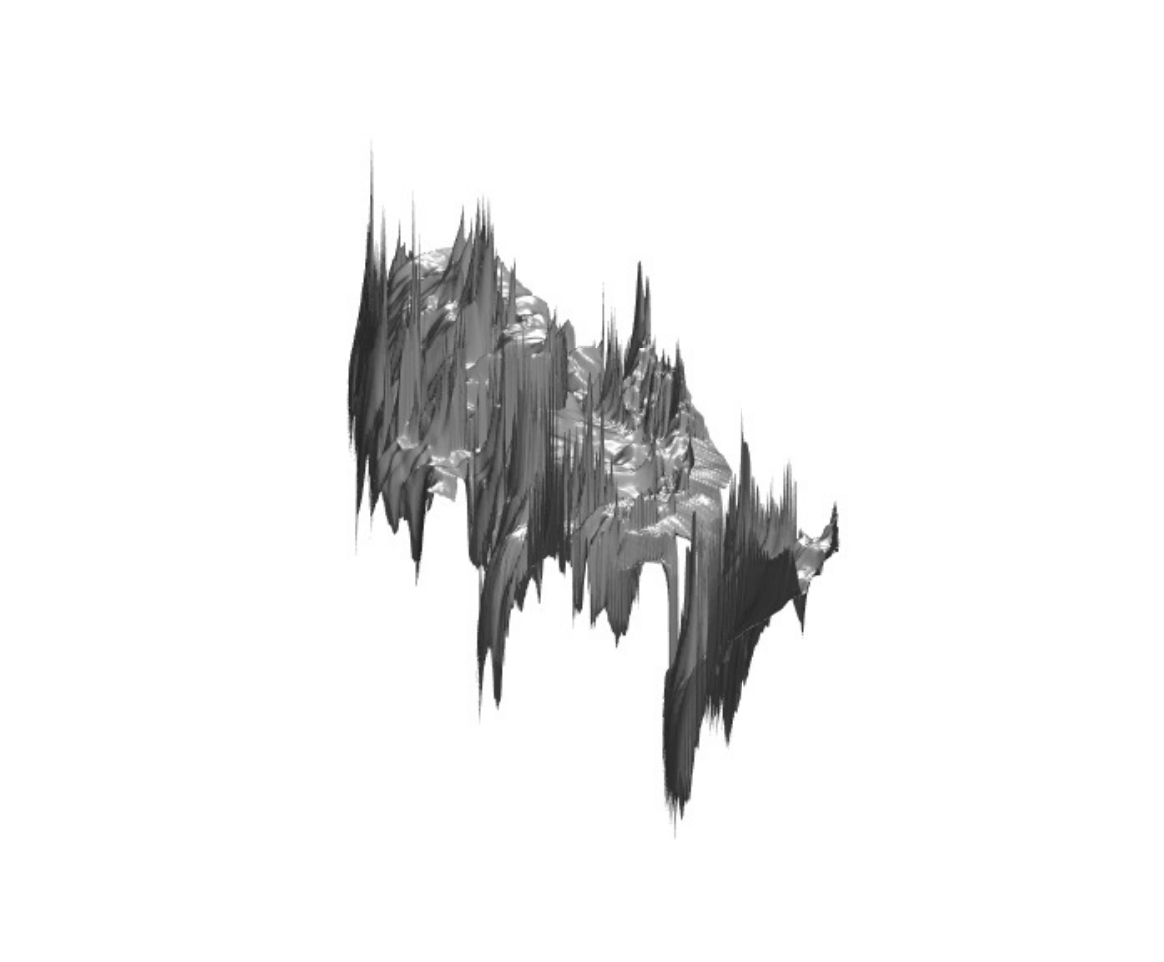} &
\includegraphics[height=0.17\linewidth,trim = 8em 9em 8em 5em,clip]{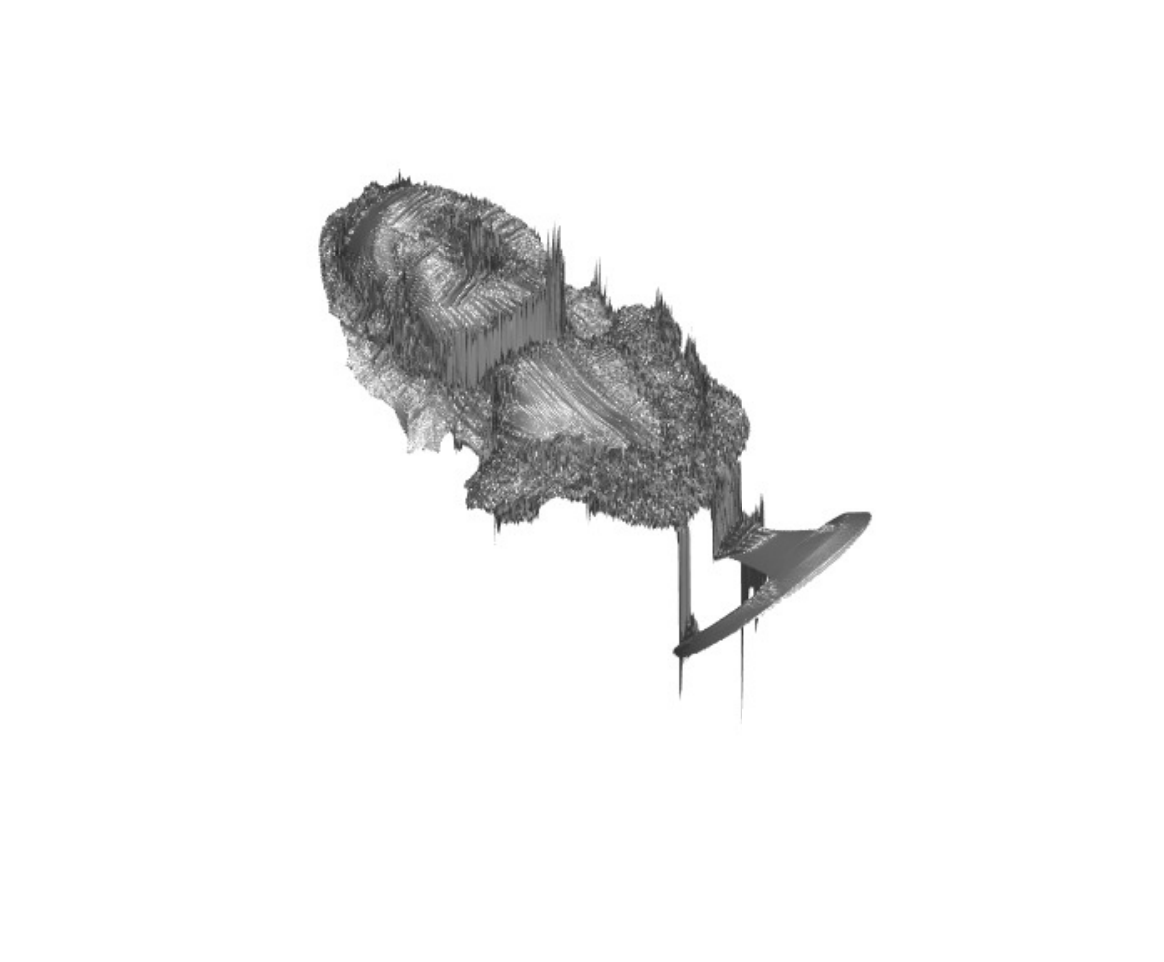} & 
\includegraphics[height=0.17\linewidth,trim = 7em 6em 7em 4.5em,clip]{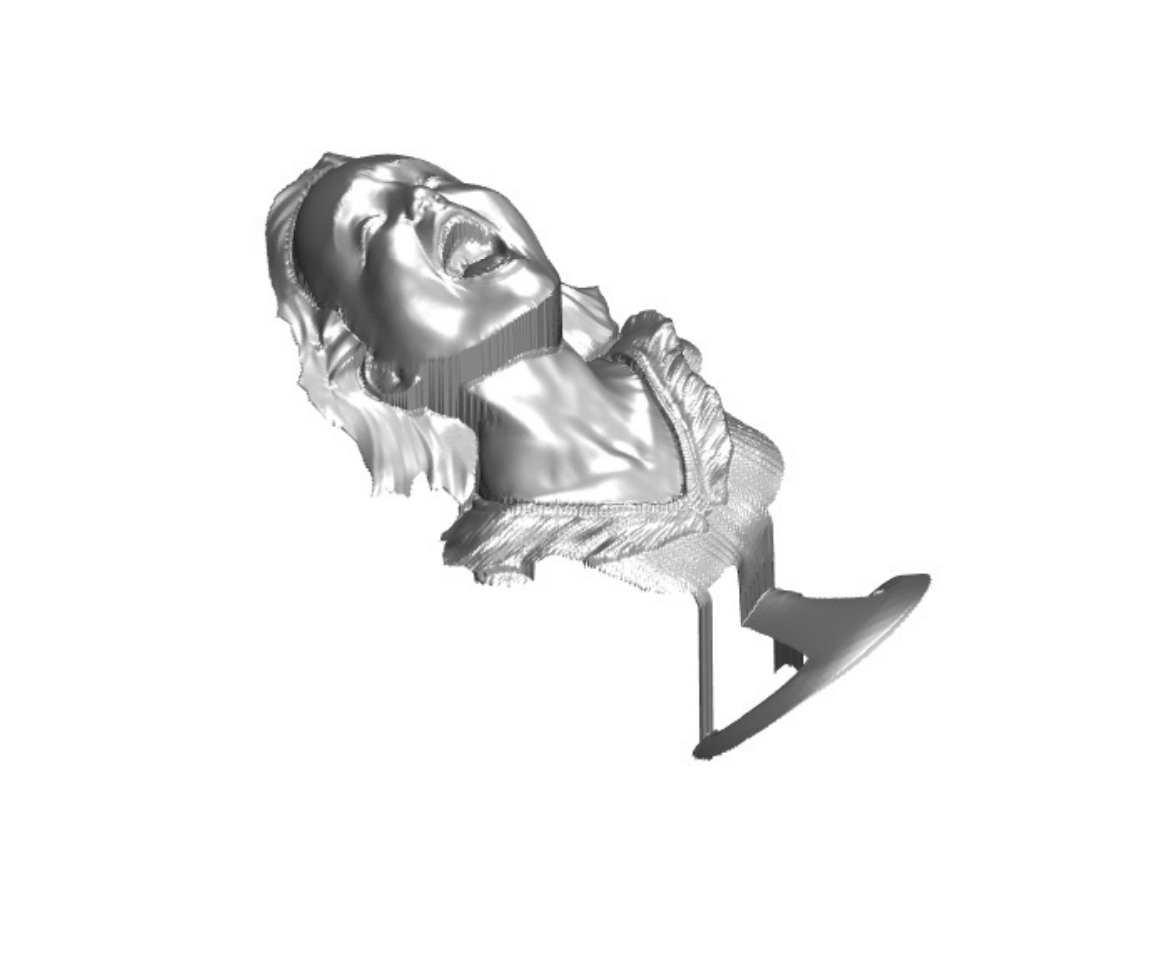} \\
{\small Input synthetic image} & {\small Fixed point~\cite{Or-El2015}} & {\small SIRFS~\cite{Barron2015}} &  {\small Proposed ADMM 3D-reconstruction} \\ 
{\small and illumination} & {\small without regularization} & {using only one scale} & (single-scale and regularization-free)
\end{tabular}
\caption{Greylevel shape-from-shading using first-order spherical harmonics. Linearization strategies such as the fixed point one used in~\cite{Or-El2015} induce artifacts if regularization is not employed. Similar issues arise in SIRFS~\cite{Barron2015} when the multi-scale approach is not used. On the contrary, the proposed ADMM approach provides satisfactory results without resorting to neither of these ad-hoc fixes. In the three experiments, the same initial shape was used (the realistic initialization of Figure~\ref{fig:experiments}). }
\label{fig:5}
\end{figure*}

In contrast to matching features or colors across images, photometric
techniques~\cite{Ackermann2015} such as shape-from-shading
(SFS)~\cite{Ikeuchi1981,Horn1989a} explicitly model the reflectance of
the object surface.  As a result, the brightness variations observed
in a single image provides an indication about variations in the
normal and geometry.  SFS is a classical ill-posed problem with
well-known ambiguities such as the one shown in Figure
\ref{fig:sketch}. From a single greylevel image both the
indentation (red curve) and the protrusion (blue curve) are possible
geometric configurations.  There exist two main strategies for solving
this ambiguity~\cite{Durou2008,Zhang1999}. Variational
methods~\cite{Horn1986} employ regularization. As a
result, they provide an approximate SFS solution which is often
over-smoothed. Alternatively, methods based on the exact resolution of
a nonlinear PDE~\cite{Lions1993} yield the highest level of detail
while implicitly enforcing smoothness in the sense of viscosity
solutions. Unfortunately, these PDE solutions lack robustness and they
require a boundary condition. Since most shape-from-shading methods
require a highly controlled illumination, they often fail when
deployed in real-world conditions outside the lab. As shown in
Figure~\ref{fig:5}, existing methods for shape-from-shading under
natural illumination~\cite{Barron2015,Or-El2015} strongly depend on
the use of a regularization mechanism, which limits their accuracy.

\paragraph{Shading-based geometry refinement. }

Obviously the mentioned concave$/$convex ambiguity disappears when using
more than one observation -- see Figure~\ref{fig:sketch}, right side.
The natural question is therefore how to combine multi-view
reconstruction with the concept of shape-from-shading.  This has long
been identified as a promising track~\cite{Blake1985}, and theoretical
guarantees on uniqueness exist~\cite{Chambolle1994}. Still, there is a
lack of practical multi-view shape-from-shading methods. Jin \etal
presented in~\cite{Jin2008} a variational solution, which relies on
regularization and may thus miss thin structures.
Besides, this solution assumes a single, infinitely distant light source and thus
cannot be applied under natural illumination. Methods combining stereo
and shading information have also been
developed~\cite{Galliani2016,Kim2016,Langguth2016,Maurer2016,Nehab2005,Samaras2000,Wu2011,Zollhoefer2015}.
Yet, they do not fully exploit the potential of shading, because they
all consider photometry as a way to refine multi-view
3D-reconstruction, which remains the baseline of the process.

\subsection{Contribution}
In this work, we revisit the challenge of multi-view
shape-from-shading.  Instead of considering SFS as a post-processing
for fine-scale geometric refinement, we rather place it at the core of
the multi-view 3D-reconstruction process. The key idea is to model the
brightness variations of each color channel and each image by means of a partial differential equation and to subsequently
couple these PDE solutions across all images and channels by means of
a variational approach.  Furthermore, rather than alternatingly
solving for shape-from-shading and concistency across all images
(which is known to lead to suboptimal solutions of poor quality), we
make use of an efficient ADMM algorithm in order to solve the
nonlinearly coupled optimization problem.  In numerous experiments we
demonstrate that the proposed variational fusion of shape-from-shading
across multiple views gives rise to highly accurate dense
reconstructions of real-world objects without the need for dense
correspondence.  We believe that the proposed extension of SFS to
multiple views will help to finally bring SFS strategies from the lab
into the real world.


\subsection{Problem Statement and Paper Organization}

Given a set of $N$ input images $I_i,~i \in \{1,\dots,N\}$ and the
reflectance function $\mathcal{R}$, we ultimately wish to estimate $N$
depth maps $z_i,~i \in \{1,\dots,N\}$, which are both consistent with
the observed images (photometric constraint), and consistent with each other (geometric constraint). The proposed framework is of the
variational form, and can be written as follows:
\begin{equation}
\underset{\{z_i\}_i}{\operatorname{min}~} \sum_{i=1}^n \mathcal{P}(\mathcal{R}(z_i) - I_i) + \mathop{\sum\sum}_{1 \leq i < j \leq N} \mathcal{G}(z_i,z_j),
\label{eq:MV_SFS} 
\end{equation}
where the photometric energy $\mathcal{P}$ and the geometric one $\mathcal{G}$ have to be chosen appropriately in order to ensure that: i) the finest details are being captured; ii) natural illumination can be considered; iii) the solution is not over-smoothed; iv) the $N$ depth maps are consistent.

The choice of the photometric energy $\mathcal{P}$ is first discussed in detail in Section~\ref{sec:SFS}. It introduces a new approach to SFS under
natural illumination which is both variational and PDE-based. It captures the finest details of a surface by avoiding regularization. Yet, since we also avoid using any boundary condition, 3D-reconstruction remains ambiguous if no initial estimate is available. To tackle this issue, we show in
Section~\ref{sec:disambiguation} that sparse correspondences
across multi-view images disambiguate the problem.

\section{Variational SFS Under Natural Illumination}
\label{sec:SFS}

This section introduces an algorithm for solving SFS under general lighting, modeled by channel-dependent, second-order spherical harmonics. We make the same assumptions as in~\cite{Johnson2011} \ie, the lighting and the albedo of the surface are known. In practice, this means that a calibration object (\eg, a sphere) with known geometry and same albedo as the surface to reconstruct must be inserted in the scene. These assumptions are usual in the SFS literature. They could be relaxed by simultaneously estimating shape, illumination and reflectance~\cite{Barron2015}, but we leave this as future work. Instead, we wish to solve SFS without resorting to any prior except differentiability of the depth map.   

 Our approach relies on the new differential SFS model~\eqref{eq:4}. To solve it in practice, we introduce the variational reformulation~\eqref{eq:8}, which separates the difficulties due to nonlinearity from those due to the non-local nature of the problem. Experimental validation is eventually conducted through an application to  shading-based depth refinement.

\subsection{Image Formation Model and Related Work}

Let $I:\,\Omega \subset \RR^2 \to \RR^C,\,(x,y) \mapsto I(x,y) = \left[I^1(x,y),\dots,I^C(x,y)\right]^\top$, be a greylevel ($C=1$) or multi-channel ($C>1$) image of a surface, where $\Omega$ represents a ``mask'' of the object being pictured.  

We assume that the surface is Lambertian, so its reflectance is characterized by the albedo $\rho$. We further consider a second-order spherical harmonics model~\cite{Basri2003,Ramamoorthi2001} for the lighting $\bl$. To deal with the spectral dependencies of reflectance and lighting, we assume both $\rho$ and $\bl$ are channel-dependent. The albedo is thus a function $\rho:\,\Omega \to \RR^C,\,(x,y) \mapsto \rho(x,y) = \left[\rho^1(x,y),\dots,\rho^C(x,y)\right]^\top$, and the lighting in each channel $c \in \{1,\dots,C\}$ is represented as a vector $\bl^c = \left[l^c_1,l^c_2,l^c_3,l^c_4,l^c_5,l^c_6,l^c_7,l^c_8,l^c_9\right]^\top \in \RR^9$. 

Eventually, let $\bn:\,\Omega \to \mathbb{S}^2 \subset \mathbb{R}^3,\,(x,y) \mapsto \bn(x,y) = \left[n_1(x,y),n_2(x,y),n_3(x,y)\right]^\top$ be the field of unit-length outward normals to the surface. 

With these notations, the image value in each channel $c \in \{1,\dots,C\}$ writes as follows, $\forall (x,y) \in \Omega$:
\begin{equation}
  I^c(x,y) = \rho^c(x,y) \, \bl^c
  \cdot 
  \begin{bmatrix}
    \mathbf{n}(x,y) \\
    1 \\
    n_1(x,y) n_2(x,y) \\ n_1(x,y) n_3(x,y) \\ n_2(x,y) n_3(x,y) \\ {n_1(x,y)}^2 - {n_2(x,y)}^2 \\ 3 {n_3(x,y)}^2 -1
  \end{bmatrix}.
  \label{eq:1} 
\end{equation}

Our goal is to recover the object shape, given its image, its albedo and the lighting. Each unit-length normal vector $\bn(x,y)$ has two degrees of freedom, thus it is in general impossible to solve Equation~\eqref{eq:1} independently in each pixel $(x,y)$. In particular, if $C=1$ and lighting is directional ($l^c_4 = \dots = l^c_9 = 0)$, Equation~\eqref{eq:1} is a single scalar equation with two unknowns. This particular situation characterizes the classic SFS problem, which is ill-posed~\cite{Horn1989a}. Its resolution  has given rise to a number of methods~\cite{Durou2008,Zhang1999}. 

Yet, few SFS methods deal with non-directional lighting. Near-field pointwise lighting has been shown to help resolving the ambiguities~\cite{Prados2005}, but only partly~\cite{Breuss2012}. Besides, to deal with more diffuse lighting such as natural outdoor illumination, spherical harmonics are better suited. First-order harmonics have been considered in~\cite{Huang2011}, but they only capture up to $90\%$ of ``real'' lighting, while this rate is over $99\%$ using second-order harmonics~\cite{Frolova2004}. 

In the context of SFS, second-order harmonics have been used in~\cite{Barron2015,Johnson2011,Richter2015}. The SFS approach of Johnson and Adelson~\cite{Johnson2011} has the same objective as ours \ie, handling multi-channel images and ``natural'' illumination, knowing the albedo and the lighting. It is shown that this general illumination model actually limits the ambiguities of SFS, since it is the intermediate case between SFS and color photometric stereo~\cite{Hernandez2007}. However, this work relies on regularization terms, which favors over-smoothed surfaces. Barron and Malik solve in~\cite{Barron2015} the more challenging problem of shape, illumination and reflectance from shading (SIRFS). By fixing the albedo and the lighting, and removing all the  regularization terms, SIRFS can be applied to SFS. However, the proposed method ``fails badly''~\cite{Barron2015} if a multi-scale strategy is not considered (see Figure~\ref{fig:5}). 
 Let us also mention for completeness the recent work in~\cite{Richter2015}, which has similar goals as ours (shape-from-shading under natural illumination), but follows an entirely different track based on discriminative learning, which requires prior training. 

Overall, there exists no purely data-driven approach to SFS under natural illumination. The rest of this section aims at filling this gap.

\subsection{Differential Model}

Since Equation~\eqref{eq:1} cannot be solved locally, it must be solved \emph{globally} over the entire domain $\Omega$. This can be achieved by assuming surface smoothness. However, in order to prevent losing the fine-scale surface details, this assumption should be as minimal as possible. In particular, regularization terms, which have been widely explored in early SFS works~\cite{Horn1986,Ikeuchi1981}, may over-smooth the solution.
Instead of having the normal vectors as unknowns and penalizing their variations, as achieved for instance in~\cite{Johnson2011}, we rather directly estimate the underlying depth map. To this end, we resort to a differential approach building upon PDEs~\cite{Lions1993}. This has the advantage of implicitly enforcing differentiability without requiring any regularization term. Let us thus first rewrite~\eqref{eq:1} as a PDE. 

Let the shape be represented as a function $z:\,\Omega \to \RR$, which is the depth map under orthographic projection, and the $\log$ of the depth map under perspective projection. In both cases, the normal to the surface is given by
\begin{equation}
  \bn = \dfrac{1}{d(z_x,z_y)} \begin{bmatrix} f z_x \\ f z_y \\ -1 - \tilde{x} z_x - \tilde{y} z_y \end{bmatrix},
  \label{eq:2}
\end{equation} 
where: $\nabla z = \left[z_x,z_y\right]^\top$ is the gradient of $z$; $(f,\tilde{x},\tilde{y}) =(1,0,0)$ under orthographic projection while, under perspective projection, $f$ is the focal length and $(\tilde{x},\tilde{y}) = (x-x_0,y-y_0)$, with $(x_0,y_0)$ the coordinates of the principal point; and $d(z_x,z_y)$ is a coefficient of normalization:
\begin{equation}
d(z_x,z_y) = \sqrt{(f {z_x})^2 + (f {z_y})^2 + \left( 1 + \tilde{x} z_x + \tilde{y} z_y \right)^2 }.
\label{eq:3}
\end{equation}
Plugging~\eqref{eq:2} into~\eqref{eq:1}, we obtain, $\forall c \in \{1,\dots,C\}$, the following nonlinear PDE in the depth $z$ over $\Omega$:
\begin{equation}
\ba^c(z_x,z_y) \cdot \begin{bmatrix} z_x \\ z_y \end{bmatrix} = b^c(z_x,z_y),
\label{eq:4}
\end{equation}
with the following definitions for the fields $\ba^c(z_x,z_y):\,\Omega \to \RR^2$ and $b^c(z_x,z_y):\,\Omega \to \RR$:
\begin{align}
\ba^c(z_x,z_y) & =  \dfrac{\rho^c}{d(z_x,z_y)} \begin{bmatrix}
 f\,l^c_1-\tilde{x}\,l^c_3 \\
 f\,l^c_2-\tilde{y}\,l^c_3 
\end{bmatrix}, \label{eq:5}\\
 b^c(z_x,z_y) & = I^c -\rho^c \, \begin{bmatrix} l^c_3 \\ l^c_4 \\ l^c_5 \\ l^c_6 \\ l^c_7 \\l^c_8 \\ l^c_9 \end{bmatrix} \cdot \begin{bmatrix}
  \frac{-1}{d(z_x,z_y)} \\
  1 \\
  \frac{f^2 z_x z_y}{d(z_x,z_y)^2} \\
  \frac{f z_x\left(-1-\tilde{x} z_x-\tilde{y} z_y\right)}{d(z_x,z_y)^2} \\
  \frac{f z_y\left(-1-\tilde{x} z_x-\tilde{y} z_y\right)}{d(z_x,z_y)^2} \\  
  \frac{f^2\left({z_x}^2-{z_y}^2\right)}{d(z_x,z_y)^2} \\  
  \frac{3\left(-1-\tilde{x} z_x -\tilde{y} z_y\right)^2}{d(z_x,z_y)^2}-1  
\end{bmatrix}.\label{eq:6}
\end{align}

Various methods have been suggested for solving PDEs akin to~\eqref{eq:4}, in some specific cases. When $C=1$, and lighting is directional and frontal (\ie, $l_3$ is the only non-zero lighting component), then~\eqref{eq:4} becomes the \emph{eikonal equation}, which was first exhibited for SFS in~\cite{Bruss1982}. After this inverse problem has caught the attention of several mathematicians~\cite{Lions1993,Rouy1992}, efficient numerical methods for approximating solutions to this well-known equation have been suggested, using for instance semi-Lagrangian schemes~\cite{Cristiani2007}. Under perspective projection, an eikonal-like equation also arises~\cite{Prados2003,Tankus2003}. The case where lighting is depth-dependent (so-called attenuation factor) is also interesting as it is less ambiguous~\cite{Prados2005}. Semi-Lagrangian schemes can also be used for the resolution, see for instance~\cite{Breuss2012}. Still, most of these differential methods require a boundary condition, or at least a state constraint, which are rarely available in practice. In addition, there currently lacks a purely data-driven numerical SFS method which would handle second-order lighting and multi-channel images: \cite{Barron2015} is strongly dependent on a multi-scale strategy, and~\cite{Johnson2011} is non-differential (per-pixel surface normal estimation) and thus resorts to regularization. The variational approach discussed hereafter solves all these issues at once.  



\begin{figure*}[!ht]
\centering
\setlength{\tabcolsep}{0.1em}
 {\renewcommand{\arraystretch}{0.6}
\begin{tabular}{c|c}
   \includegraphics[height = 0.12\linewidth]{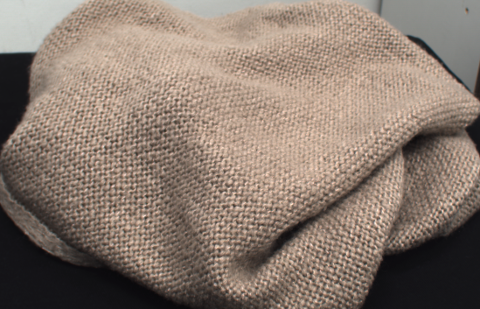} \hspace*{-2.75em}\includegraphics[height = 0.05\linewidth]{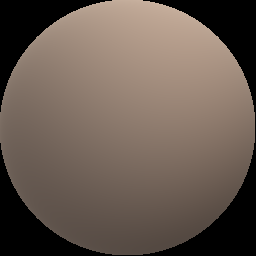}  
   \includegraphics[height = 0.12\linewidth]{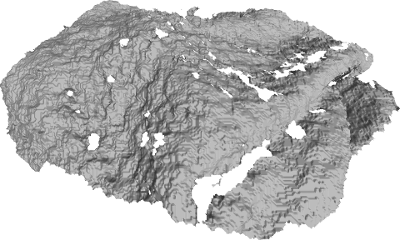}\includegraphics[height = 0.12\linewidth]{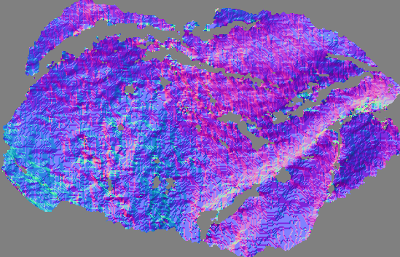} &
   \includegraphics[height = 0.12\linewidth]{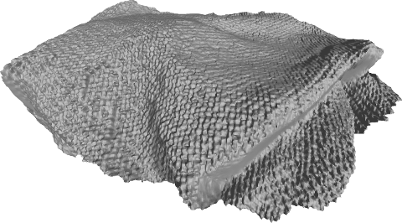}\includegraphics[height = 0.12\linewidth]{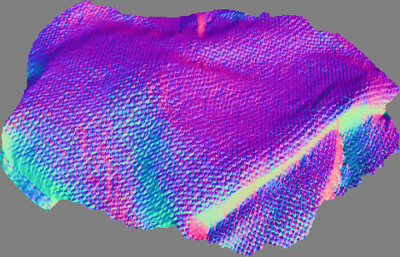} \\
   {\small A single input real image with illumination + Initial shape} & {\small Shape-from-shading without any regularization}                
 \end{tabular}
\begin{tabular}{ccc|ccc|ccc}
  \includegraphics[height=0.12\linewidth]{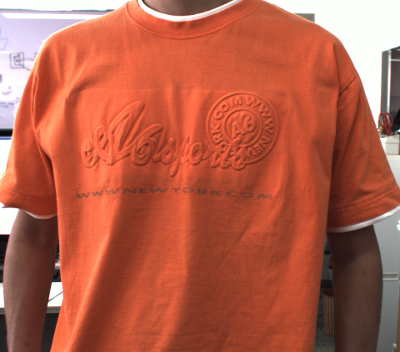}&
  \includegraphics[height=0.12\linewidth]{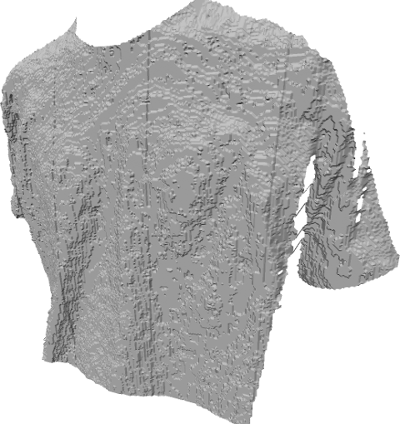} &
  \includegraphics[height=0.12\linewidth]{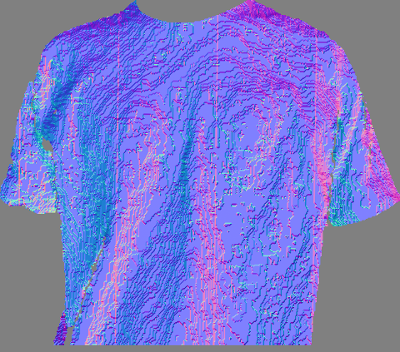} \quad & ~
  \includegraphics[height=0.12\linewidth]{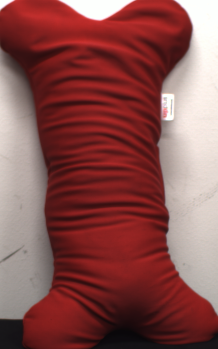}&
  \includegraphics[height=0.12\linewidth]{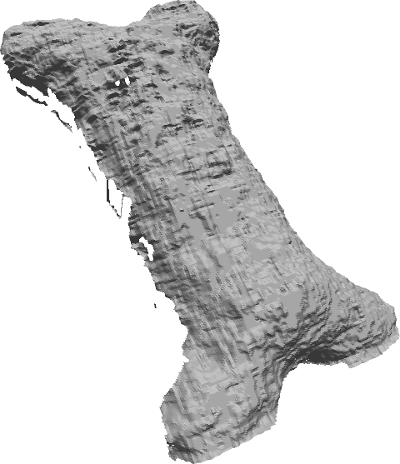} &
  \includegraphics[height=0.12\linewidth]{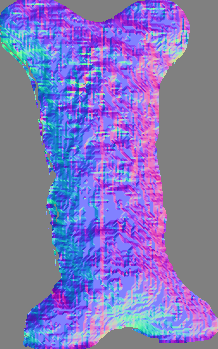} \quad & ~ 
  \includegraphics[height=0.12\linewidth]{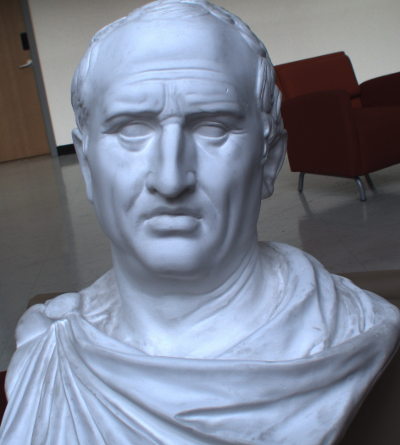}&
  \includegraphics[height=0.12\linewidth]{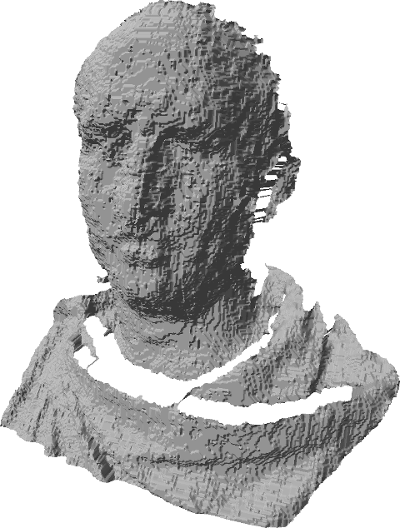} &
  \includegraphics[height=0.12\linewidth]{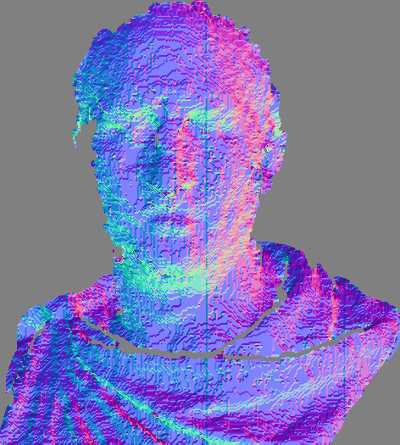} \\
  \includegraphics[height=0.08\linewidth]{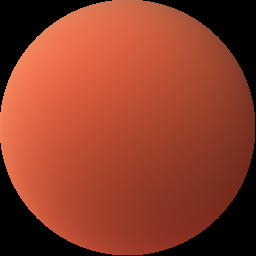}&
  \includegraphics[height=0.12\linewidth]{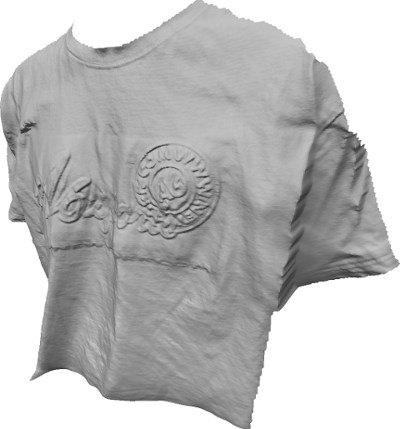} &
  \includegraphics[height=0.12\linewidth]{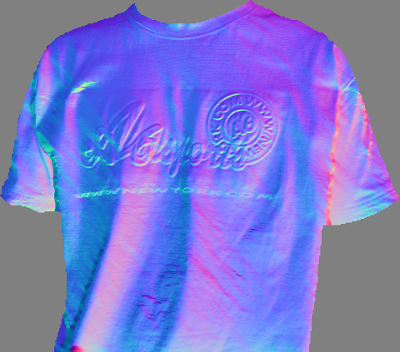} \quad & ~
  \includegraphics[height=0.08\linewidth]{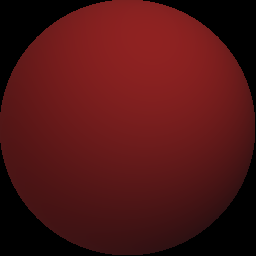}&
  \includegraphics[height=0.12\linewidth]{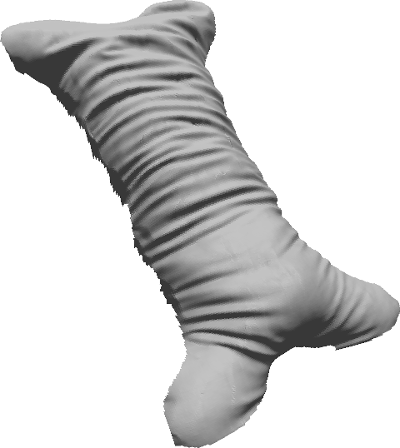} &
  \includegraphics[height=0.12\linewidth]{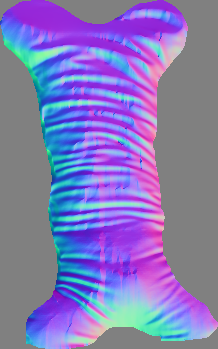} \quad & ~
  \includegraphics[height=0.08\linewidth]{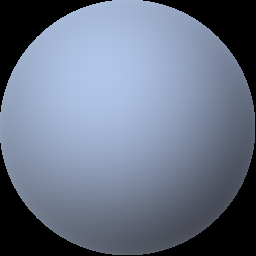}&
  \includegraphics[height=0.12\linewidth]{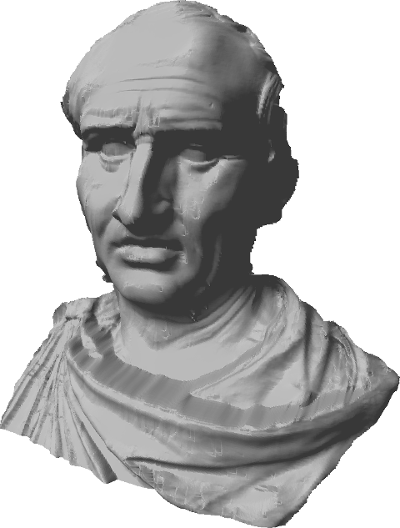} &
  \includegraphics[height=0.12\linewidth]{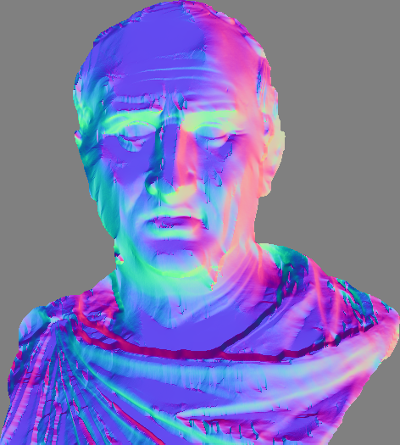} \\
  \multicolumn{6}{c}{\small \qquad \qquad\qquad\qquad\qquad\qquad\qquad\qquad\qquad\qquad\qquad Same, on three other RGB-D datasets}    
\end{tabular}}\vspace{-0.5em}
\caption{Depth refinement of RGB-D data using the proposed SFS method. }
\label{fig:RGBD}
\end{figure*}

\subsection{Variational Formulation}

The $C$ PDEs in~\eqref{eq:4} are in general incompatible due to noise. Thus, an approximate solution must be sought. For simplicity, we follow here a least-squares approach:
\begin{equation}
  \underset{\substack{z:\,\Omega \to \RR}}{\min}  \! \displaystyle\sum_{c=1}^C \!\left\| \ba^c(z_x,\!z_y) \!\cdot\! \begin{bmatrix} z_x \\ z_y \end{bmatrix} \!- b^c(z_x,\!z_y) \right\|^2_{2},  
\label{eq:7}
\end{equation}
where $\|\cdot\|_{2}$ is the $\ell^2$ norm over the domain $\Omega$.

If the fields $\ba^c$ and $b^c$ were not dependent on $z$, then~\eqref{eq:7} would be a linear least-squares problem. 
In recent works on shading-based refinement~\cite{Or-El2015}, it is suggested to proceed iteratively, by freezing these terms at each iteration.
Although this ``fixed point'' strategy looks appealing, Figure~\ref{fig:5} shows that it induces artifacts, and thus regularization must be employed~\cite{Or-El2015}. Other artifacts also arise in SIRFS~\cite{Barron2015}, when the multi-scale strategy is not employed. 

Instead of eliminating artifacts by regularization, which may induce a loss of geometric details, we rather separate the difficulty induced by the nonlinearity from that induced by the dependency on the gradient. To this end, we introduce an auxiliary variable $\theta:\,\Omega \to \RR^2$, and rewrite~\eqref{eq:7} in the following, equivalent, manner:
\begin{equation}
\begin{array}{l}
  \underset{\substack{z:\,\Omega \to \RR \\ \theta:\, \Omega \to\RR^2}}{\min~}  \displaystyle\sum_{c=1}^C \left\| \ba^c(\theta) \cdot \begin{bmatrix} z_x \\ z_y \end{bmatrix} - b^c(\theta) \right\|_2^2  \\
  \text{s.t.}~ (z_x,z_y) = \theta.
\end{array}
\label{eq:8}
\end{equation}


\begin{figure*}
  \setlength{\tabcolsep}{0.0em} 
{\renewcommand{\arraystretch}{0.6}
\begin{tabular}{ccccc}
    \multicolumn{2}{c}{\includegraphics[width=0.18\linewidth,trim = 8em 6em 7em 4em,clip]{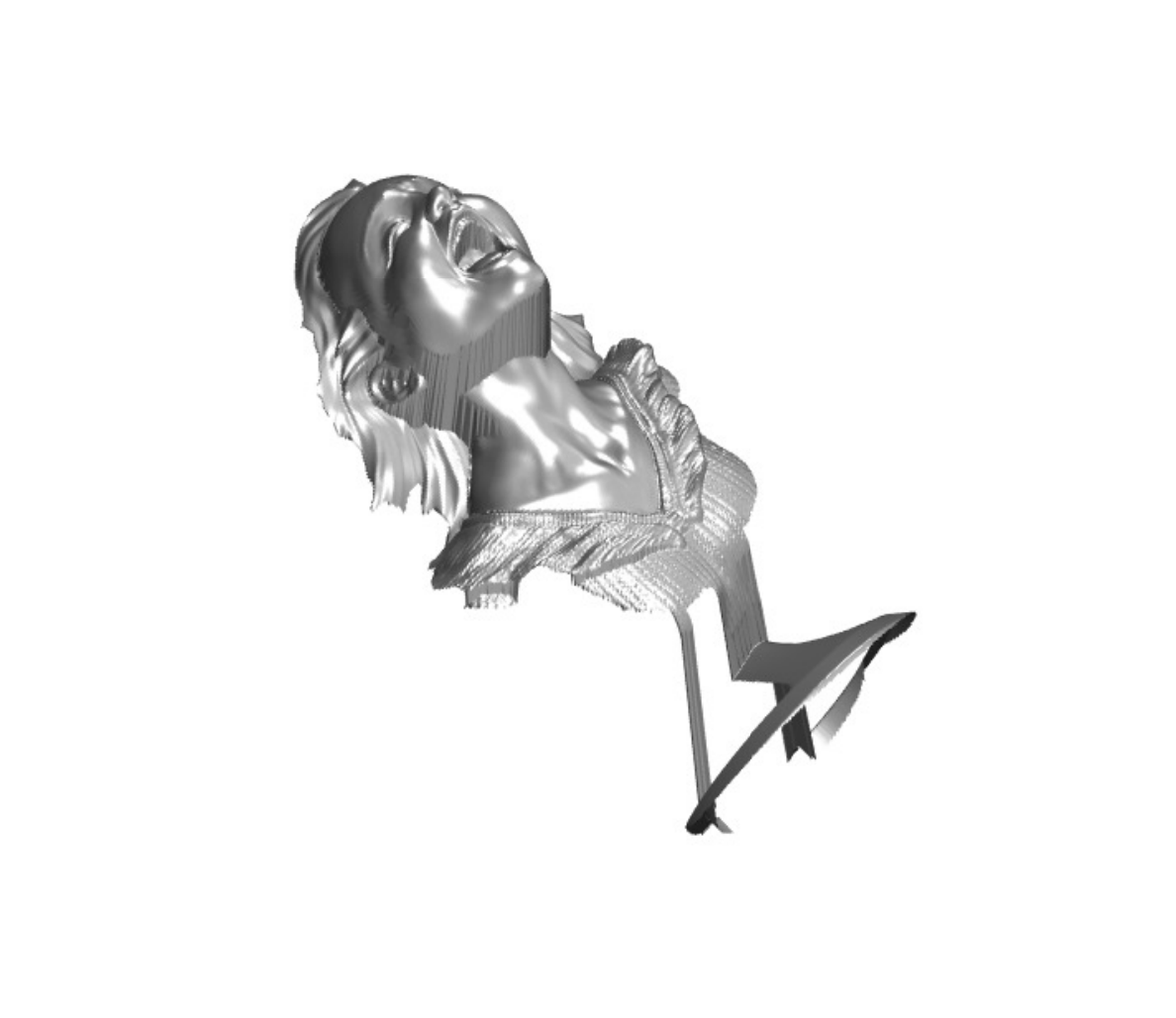}} &
    \includegraphics[height=0.18\linewidth]{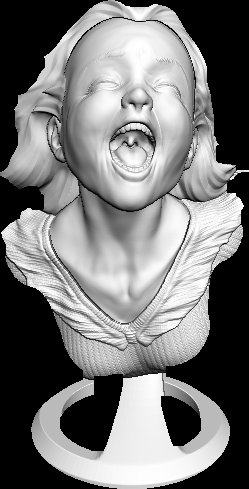}
    \includegraphics[height=0.07\linewidth]{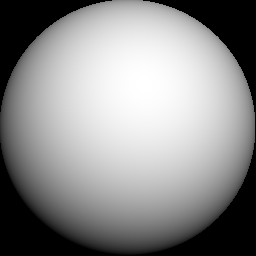} &
    \includegraphics[height=0.18\linewidth]{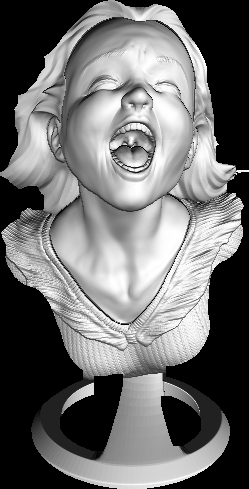} 
    \includegraphics[height=0.07\linewidth]{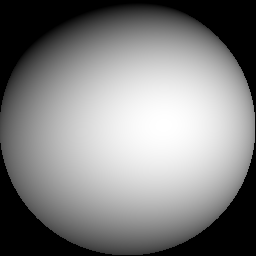} &  
    \includegraphics[height=0.18\linewidth]{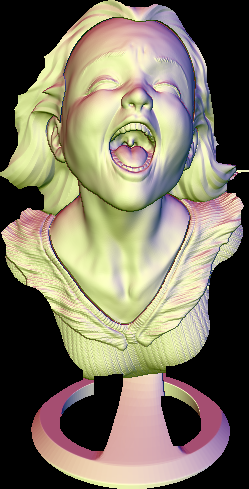} 
    \includegraphics[height=0.07\linewidth]{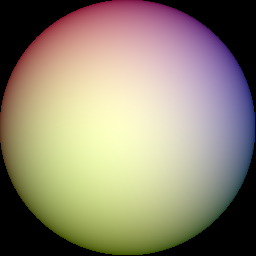} \\
 \multicolumn{2}{c}{ {\small Ground truth}} & {\small Greylevel, first-order~\eqref{eq:l1}} & {\small Greylevel, second-order~\eqref{eq:l2}} & {\small Colored, second-order~\eqref{eq:l3}} \\
  \multirow{3}{*}{
  \begin{tabular}{c}
  \includegraphics[width=0.15\linewidth,trim = 7em 6em 6em 4em,clip]{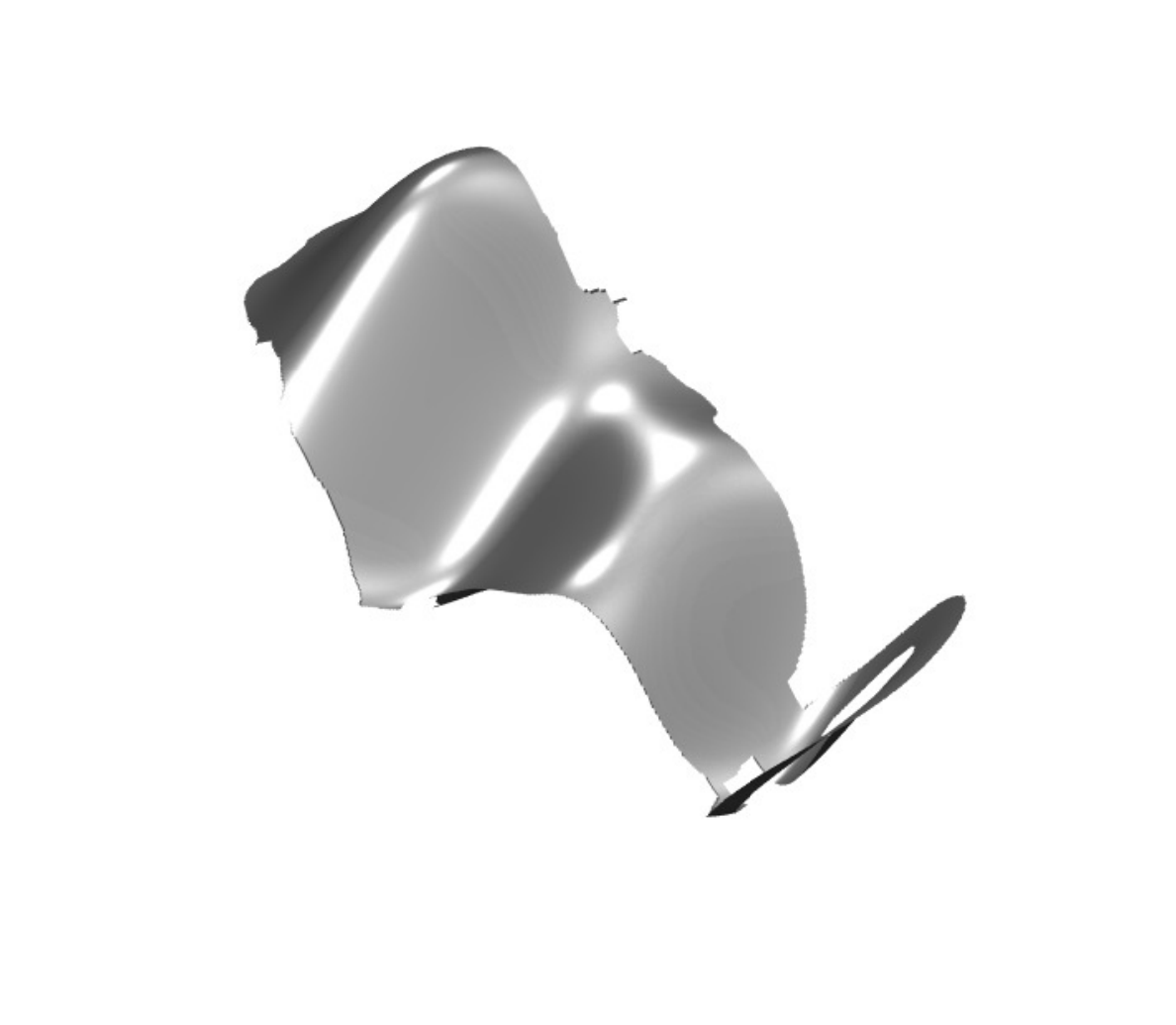} \\ Non-realistic \\ initialization 
  \end{tabular}} &
 \parbox[t]{2.5mm}{\rotatebox[origin=c]{90}{Ours}} &
 \includegraphics[height=0.15\linewidth,trim = 8em 6em 7em 4em,clip]{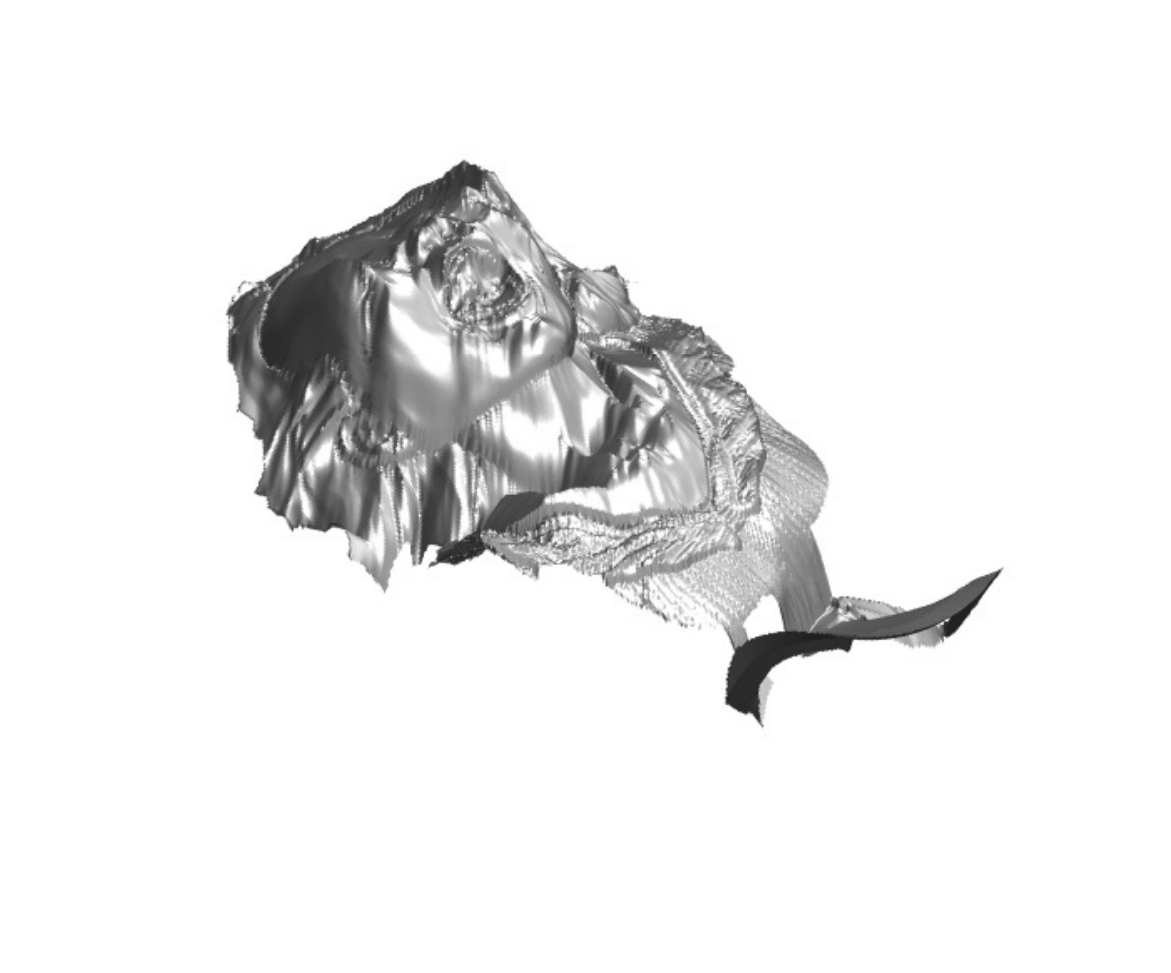}
 \includegraphics[height=0.15\linewidth]{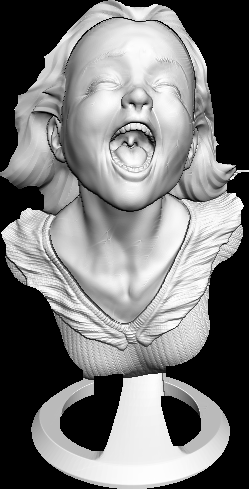} &
 \includegraphics[height=0.15\linewidth,trim = 6em 6em 5em 4em,clip]{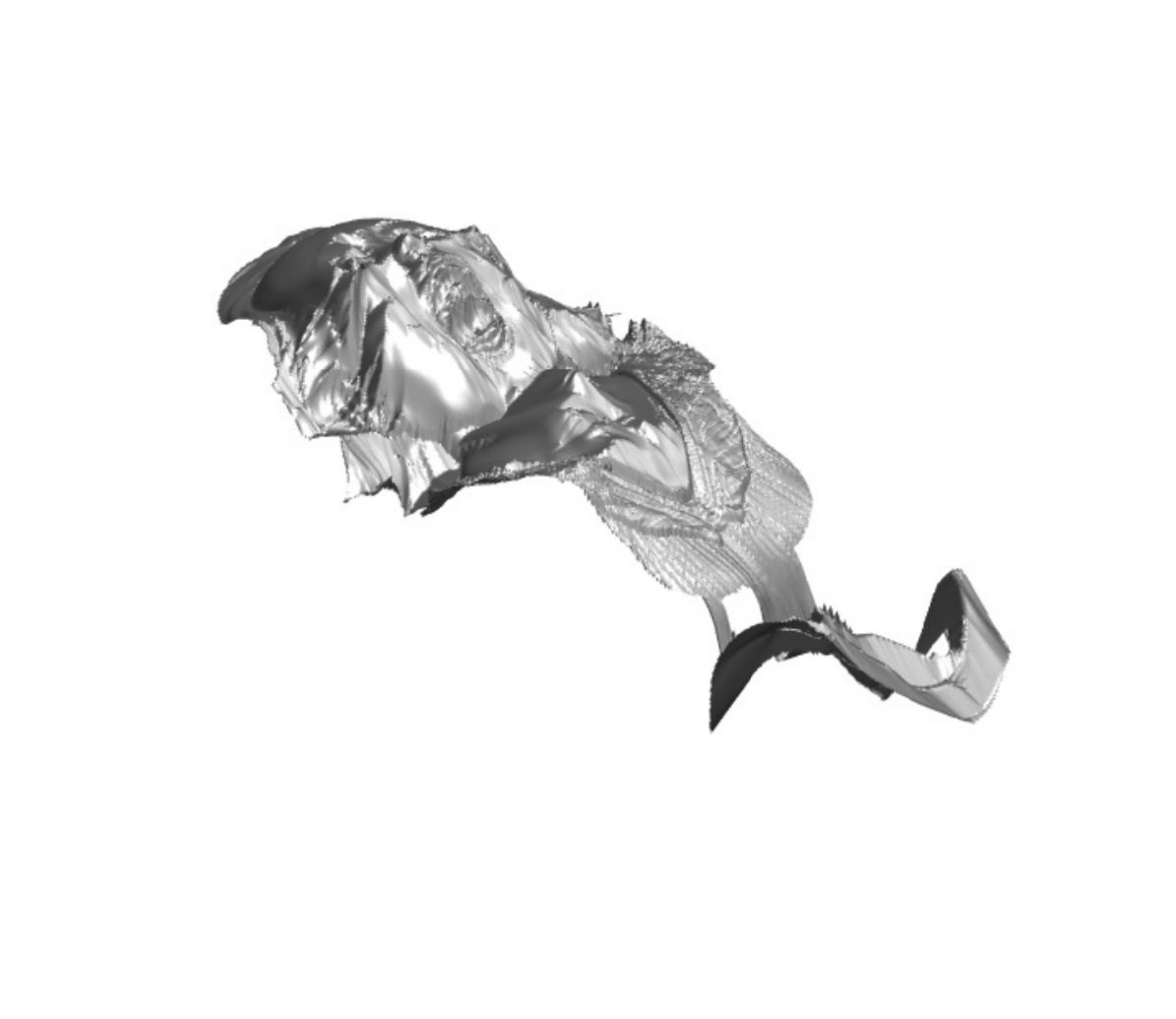} 
\includegraphics[height=0.15\linewidth]{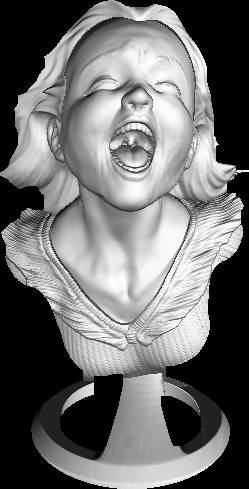} &
\includegraphics[height=0.15\linewidth,trim = 6em 6em 5em 4em,clip]{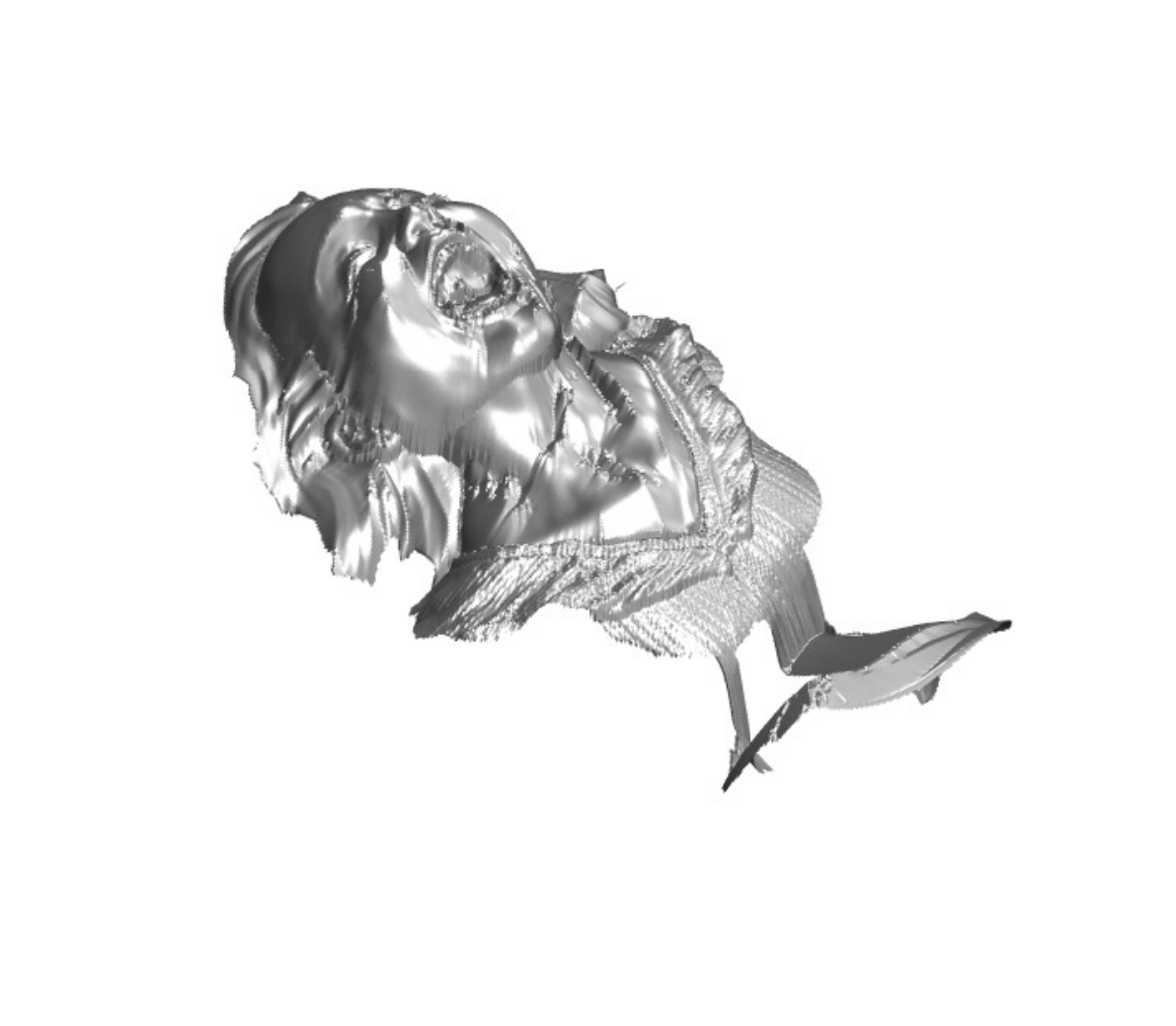} 
\includegraphics[height=0.15\linewidth]{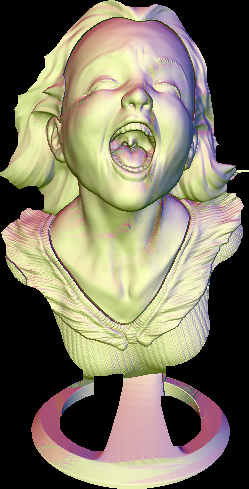} \\
 & & {\small MAE-N $= \textbf{27.91}$, RMSE-I $= \textbf{0.01}$} & {\small MAE-N $= \textbf{37.46}$, RMSE-I $= \textbf{0.02}$} & {\small MAE-N $= \textbf{13.80}$, RMSE-I $= \textbf{0.04}$} \\
 & 
 \parbox[t]{2.5mm}{\rotatebox[origin=c]{90}{SIRFS}} &
 \includegraphics[height=0.15\linewidth,trim = 6em 6em 6em 4em,clip]{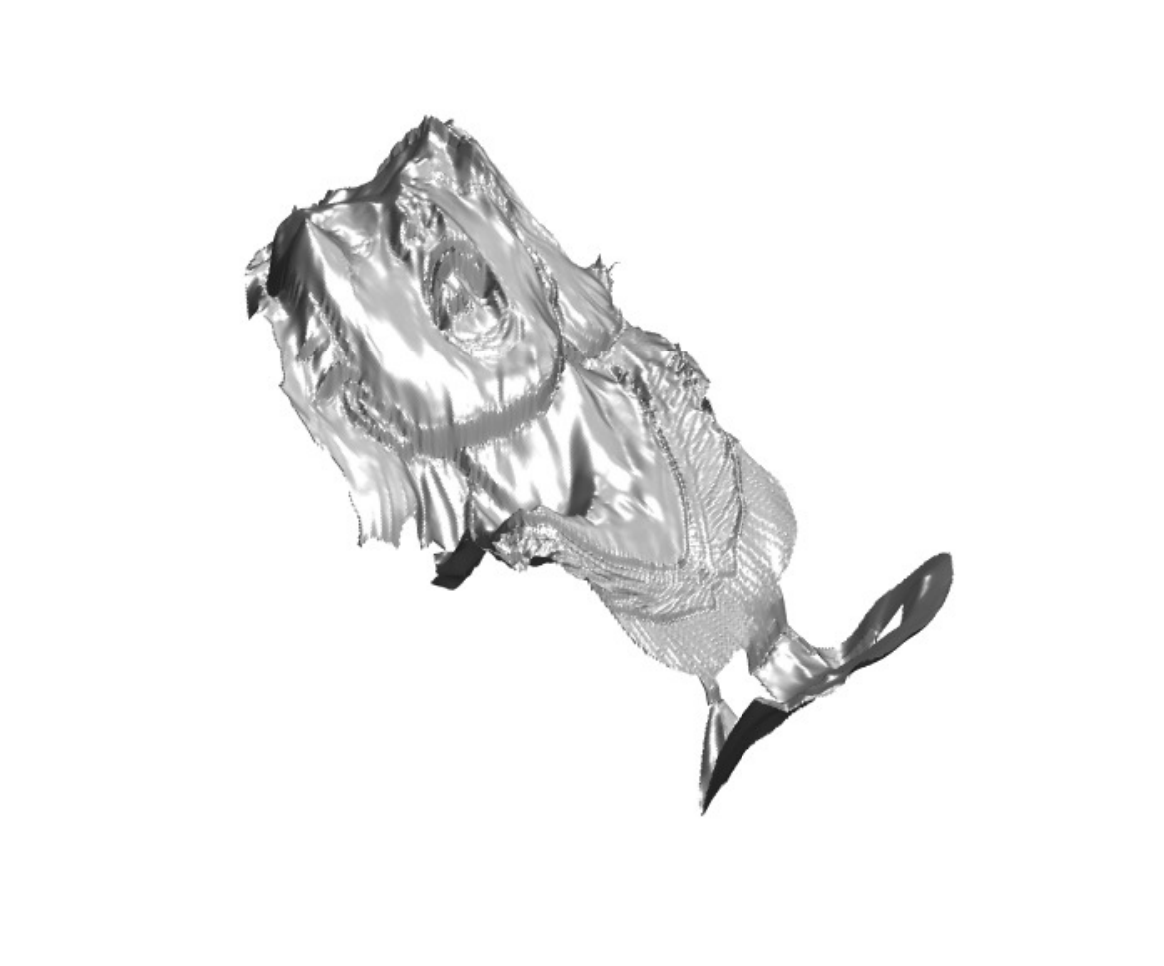}
 \includegraphics[height=0.15\linewidth]{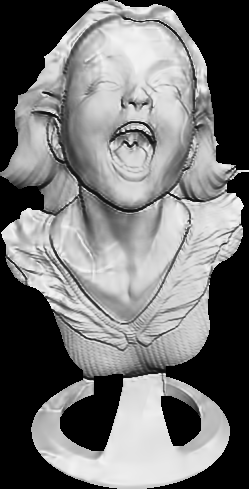} &
 \includegraphics[height=0.15\linewidth,trim = 6em 6em 6em 4em,clip]{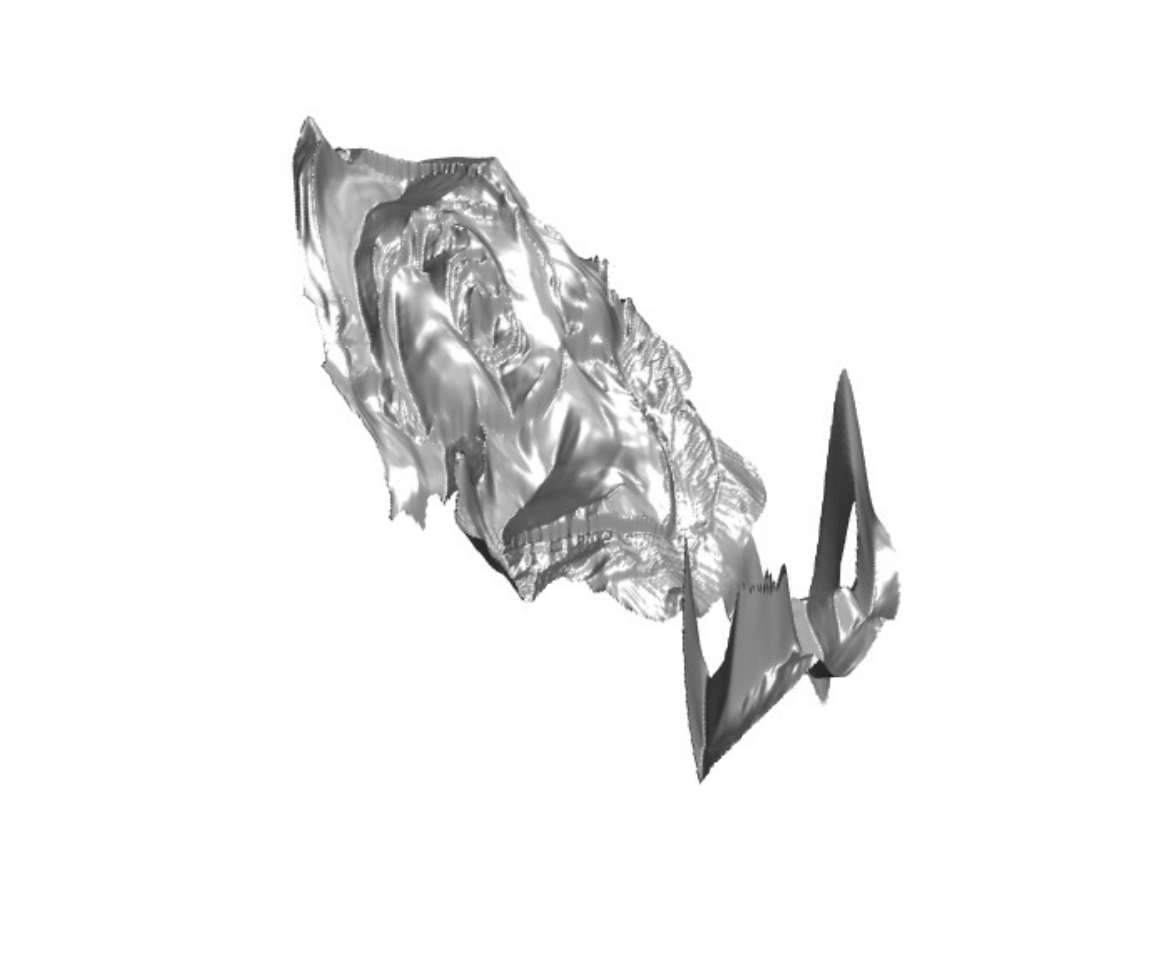} 
\includegraphics[height=0.15\linewidth]{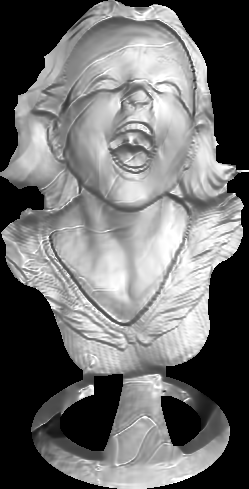} &
\includegraphics[height=0.15\linewidth,trim = 6em 6em 6em 4em,clip]{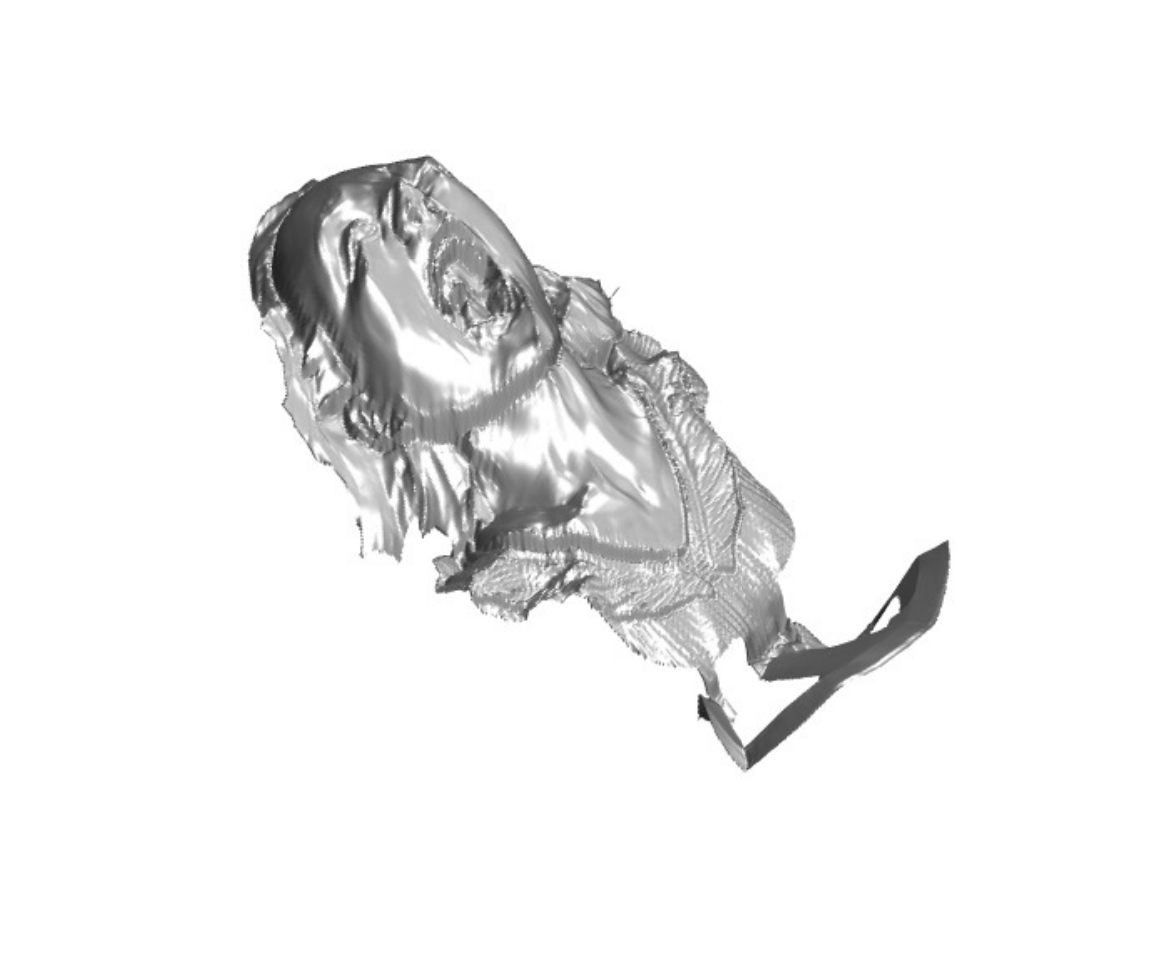} 
\includegraphics[height=0.15\linewidth]{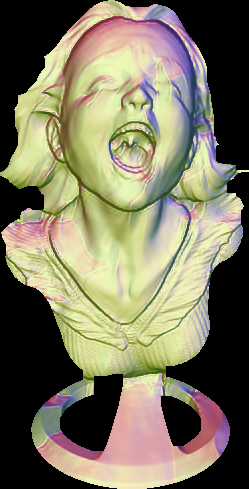} \\
 & & {\small MAE-N $= 34.90$, RMSE-I $= 0.08$} & {\small MAE-N $= 46.77$, RMSE-I $= 0.10$} & {\small MAE-N $= 24.49$, RMSE-I $= 0.08$} \\
 \multirow{2}{*}{
 \begin{tabular}{c}
  \includegraphics[width=0.15\linewidth,trim = 7em 6em 6em 4em,clip]{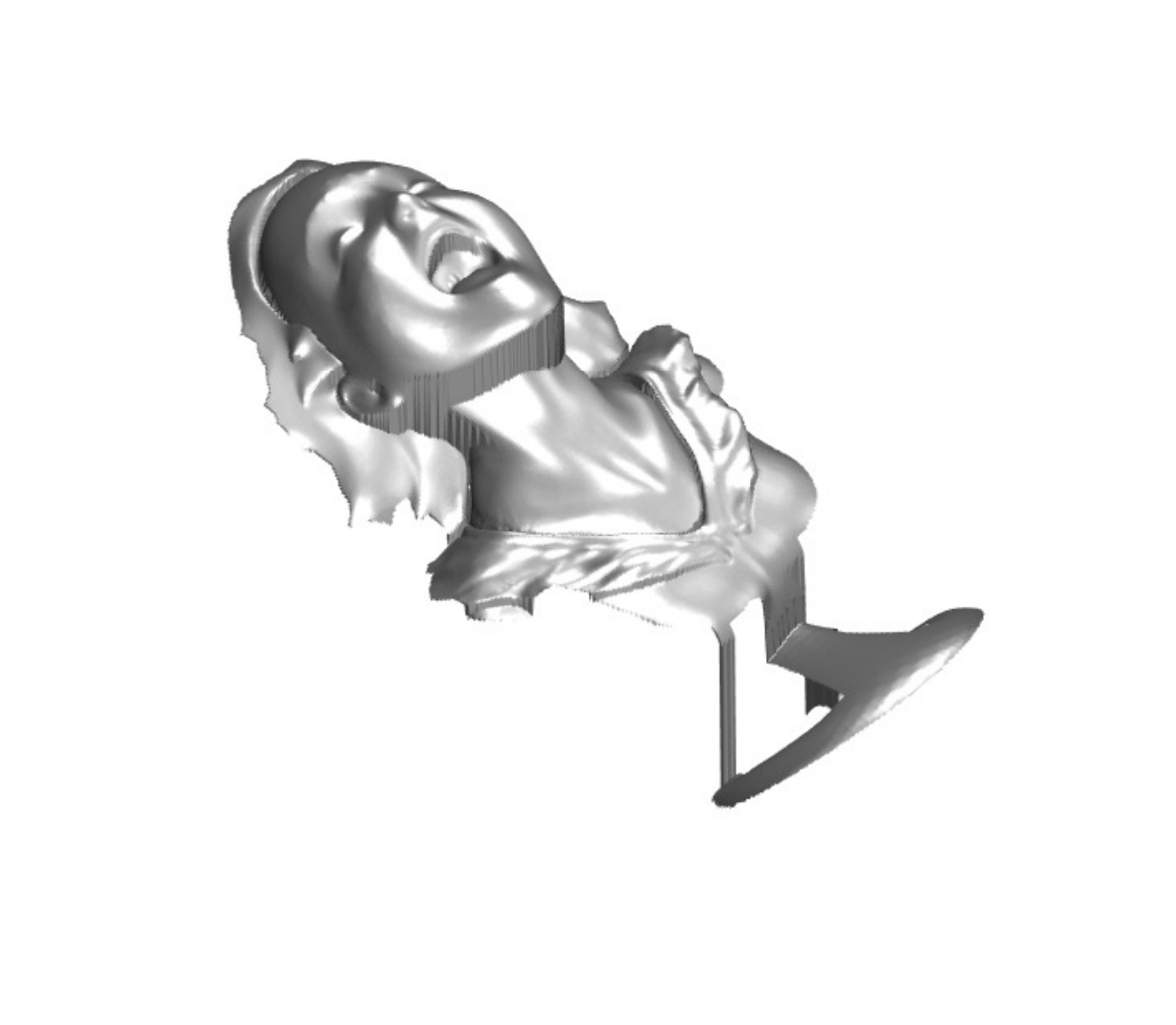} \\
  Realistic \\
   initialization \end{tabular}} &
 \parbox[t]{2.5mm}{\rotatebox[origin=c]{90}{Ours}} &    
 \includegraphics[height=0.15\linewidth,trim = 6em 6em 6em 4em,clip]{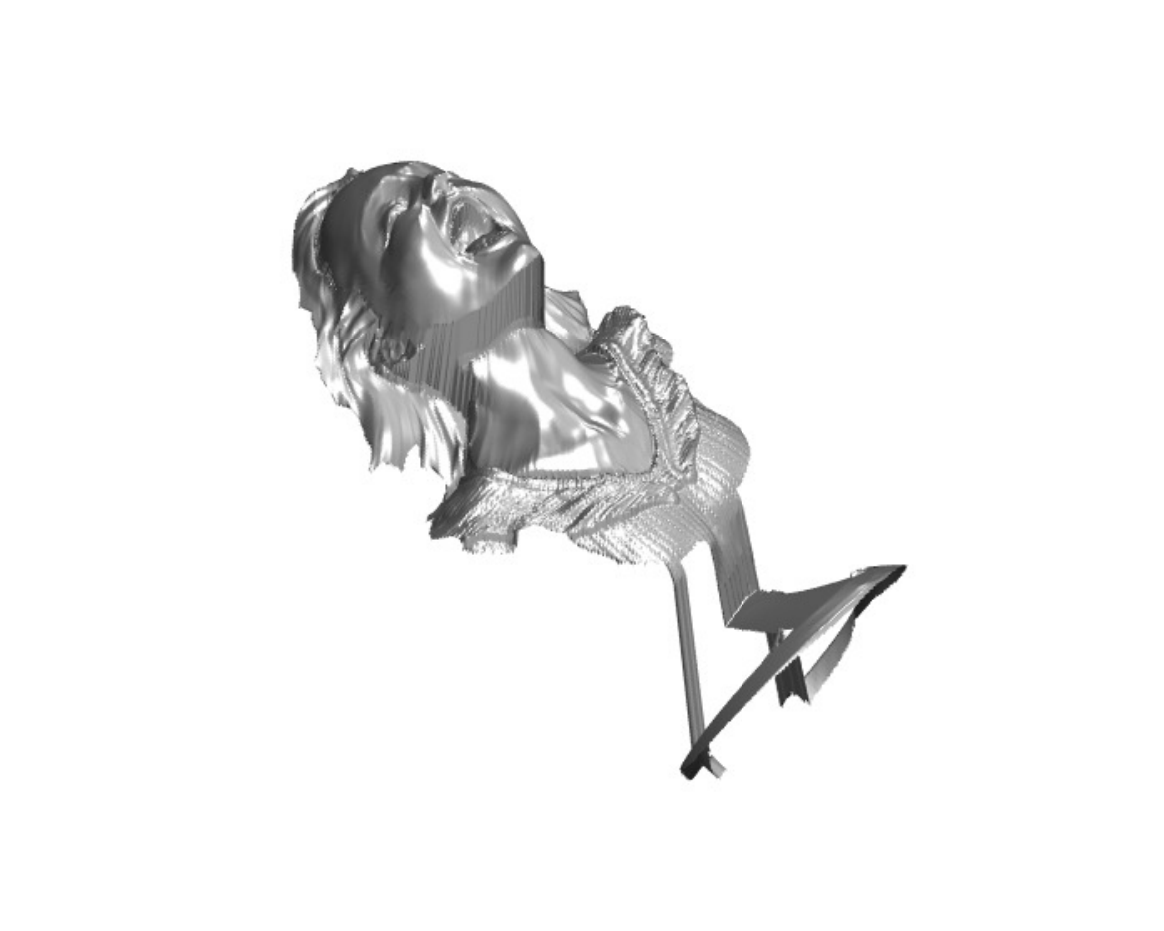}
 \includegraphics[height=0.15\linewidth]{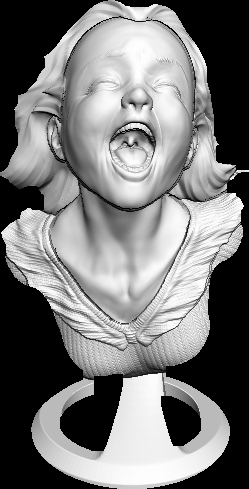} & 
 \includegraphics[height=0.15\linewidth,trim = 6em 6em 7em 4em,clip]{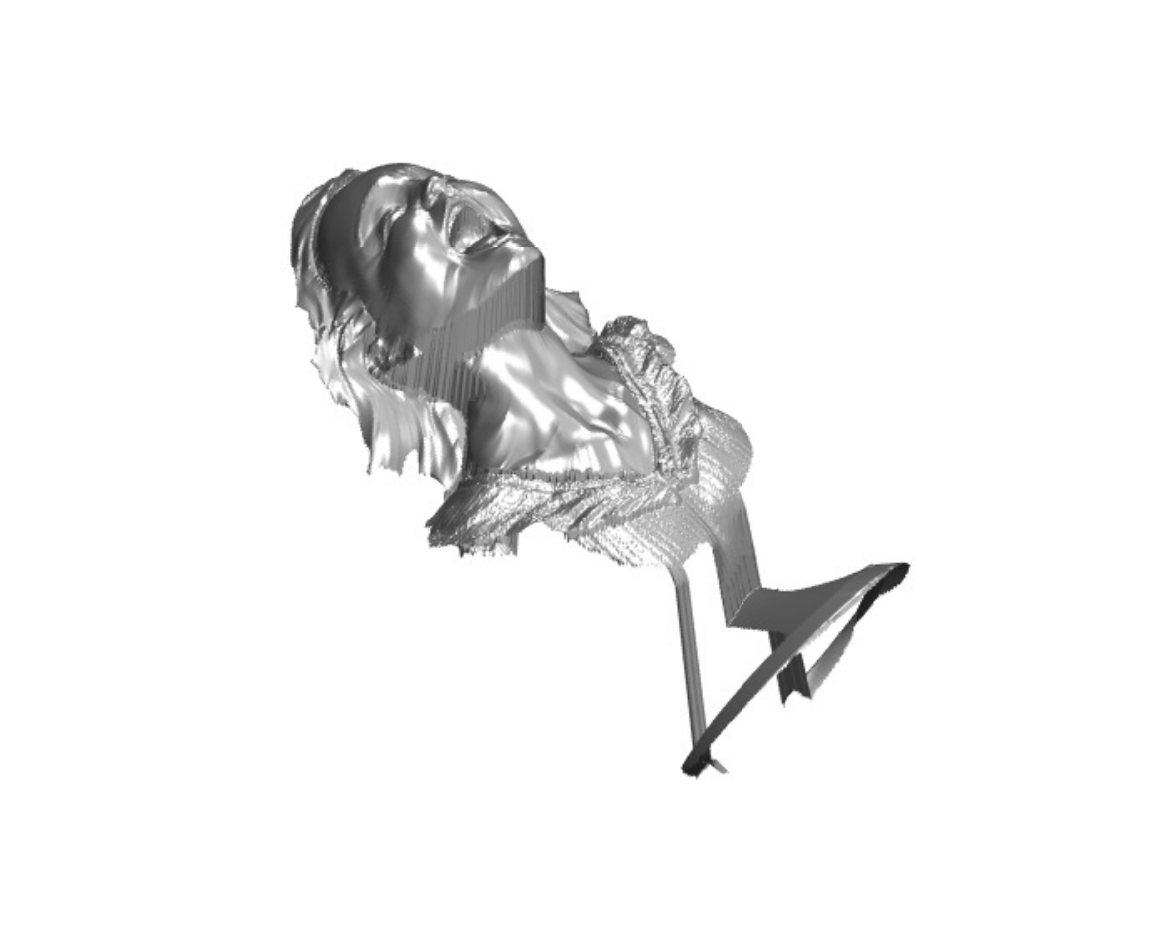} 
\includegraphics[height=0.15\linewidth]{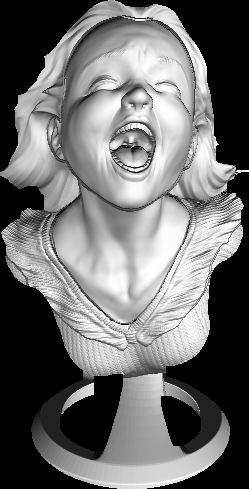} &
\includegraphics[height=0.15\linewidth,trim = 6em 6em 7em 4em,clip]{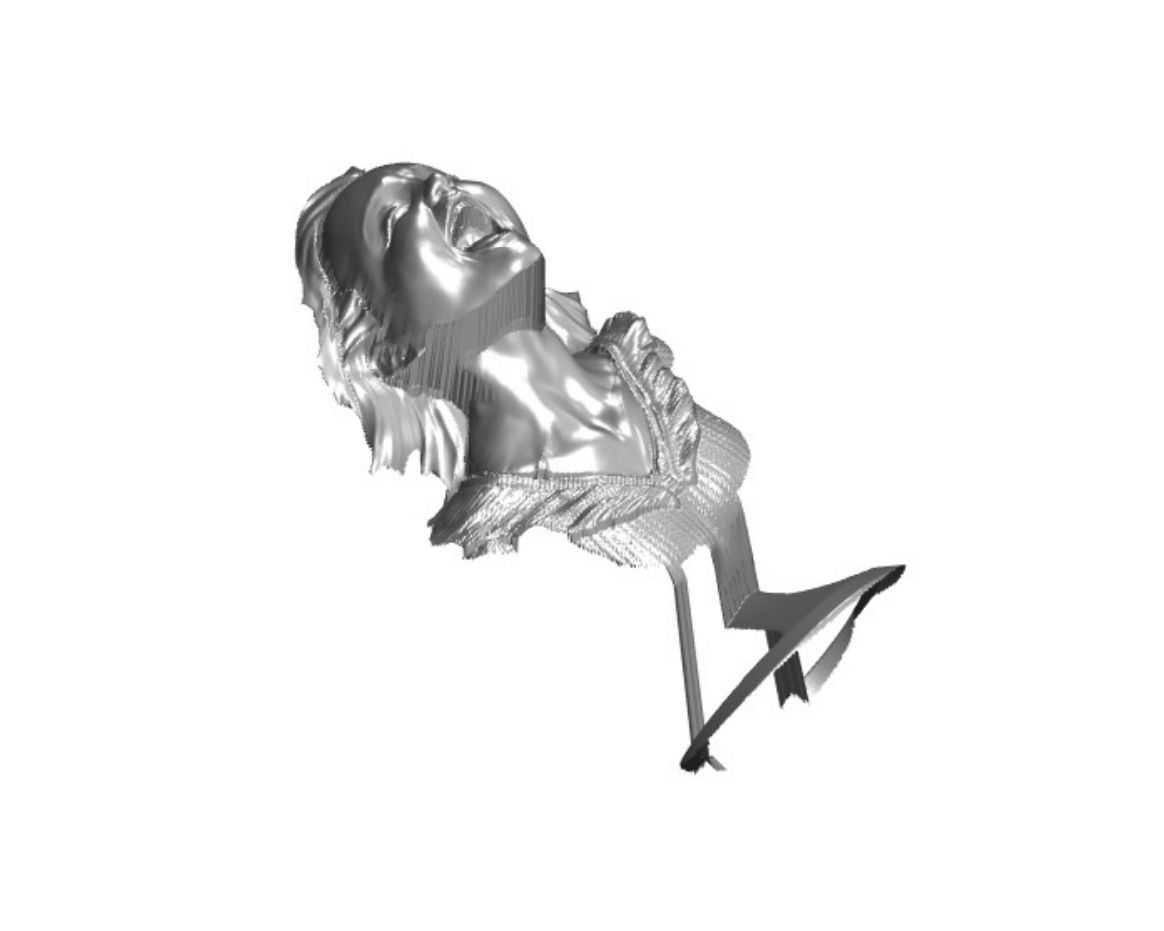} 
\includegraphics[height=0.15\linewidth]{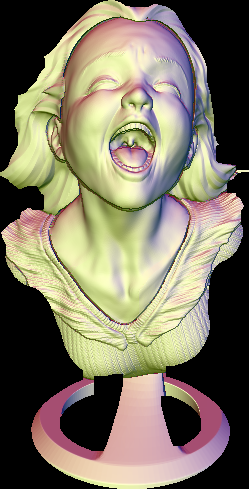} \\
 & & {\small MAE-N $= \textbf{8.74}$, RMSE-I $= \textbf{0.03}$} & {\small MAE-N $= \textbf{9.93}$, RMSE-I $= \textbf{0.04}$} & {\small MAE-N $= \textbf{3.61}$, RMSE-I $= \textbf{0.03}$} \\
  & 
 \parbox[t]{2.5mm}{\rotatebox[origin=c]{90}{SIRFS}} &  
 \includegraphics[height=0.15\linewidth,trim = 6em 6em 6em 4em,clip]{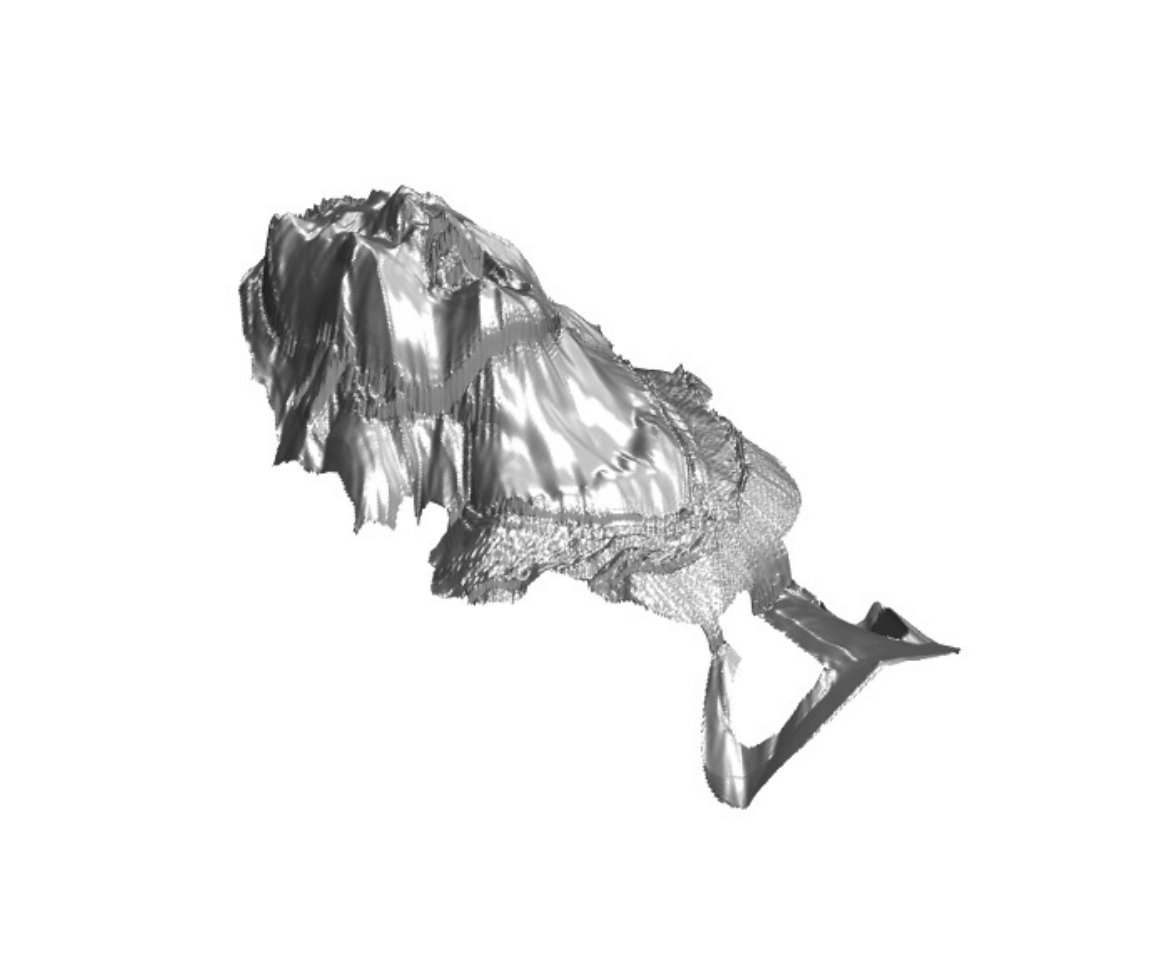}
 \includegraphics[height=0.15\linewidth]{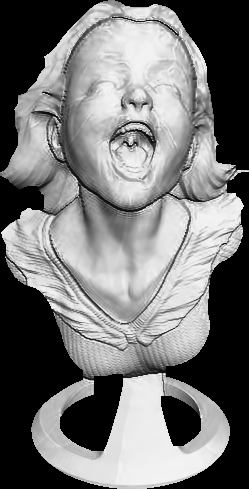} &
 \includegraphics[height=0.15\linewidth,trim = 6em 6em 6em 4em,clip]{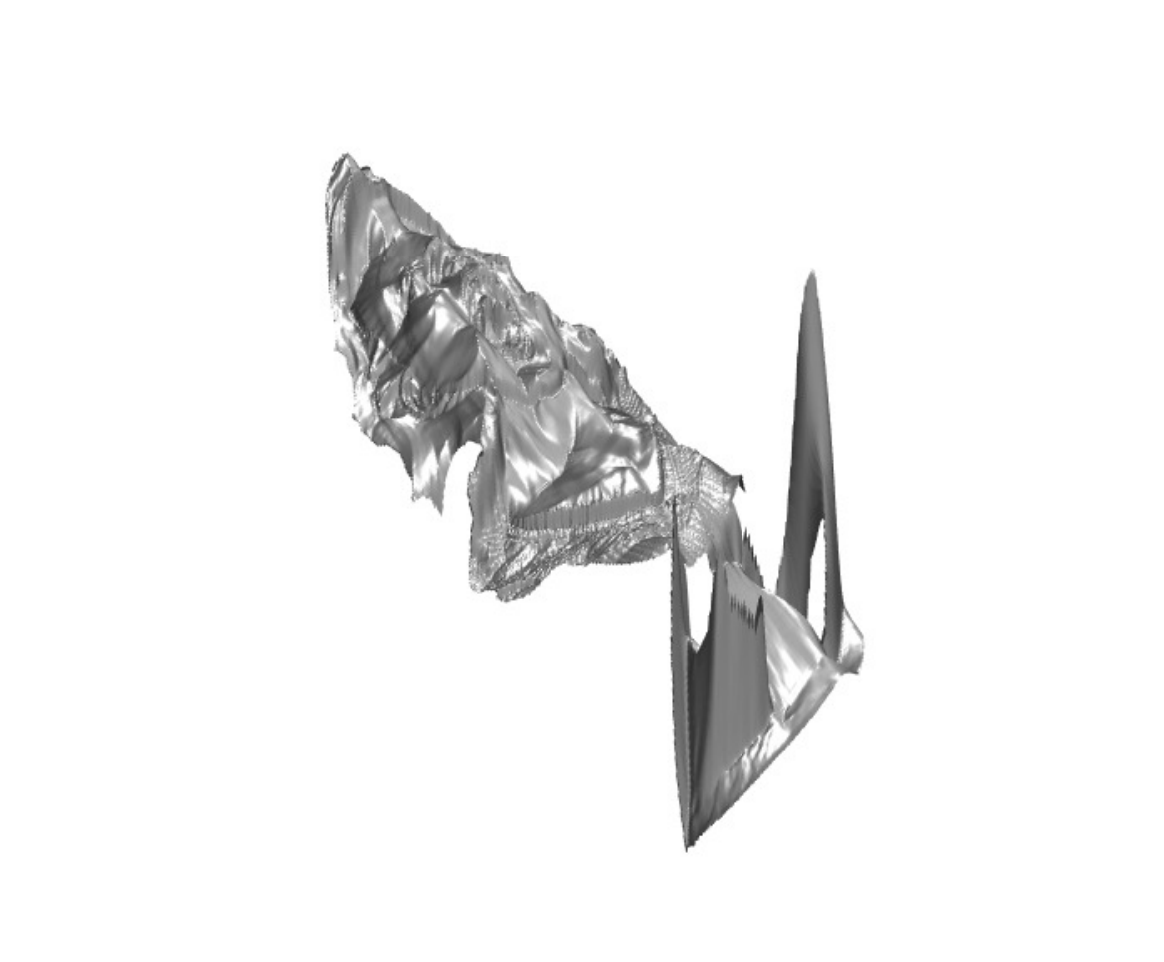} 
\includegraphics[height=0.15\linewidth]{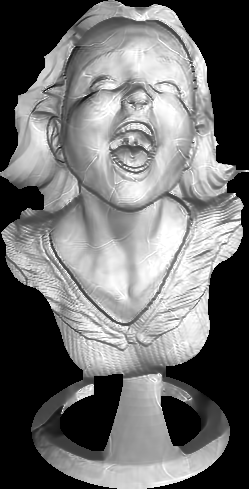} &
\includegraphics[height=0.15\linewidth,trim = 6em 6em 5em 4em,clip]{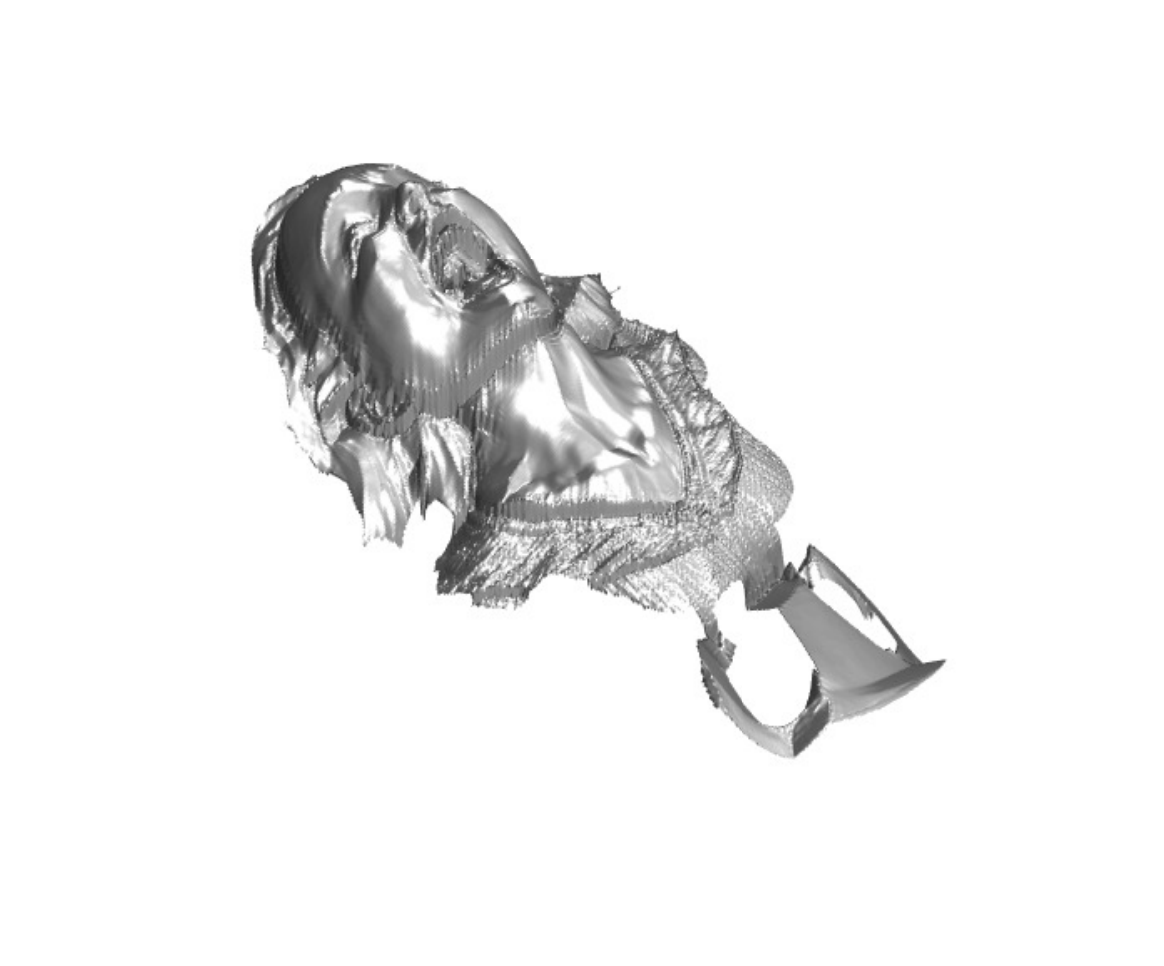} 
\includegraphics[height=0.15\linewidth]{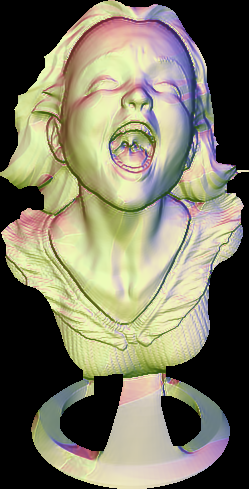} \\
 & & {\small MAE-N $= 28.46$, RMSE-I $= 0.06$} & {\small MAE-N $= 45.33$, RMSE-I $= 0.07$} & {\small MAE-N $= 23.52$, RMSE-I $= 0.07$} \\
 \\
  \end{tabular}
  }
  \caption{Evaluation of our SFS approach against the multi-scale one from SIRFS~\cite{Barron2015}, in three different lighting situations and using two different initial estimates. For each experiment, we provide the mean angular error w.r.t. ground truth normals (MAE-N, in degrees), and the root mean square error between the input synthetic image and the one simulated from the estimated depth (RMSE-I). Our method outperforms SIRFS in all tests, in terms of both metrics.} 
  \label{fig:experiments}
\end{figure*}

We then turn~\eqref{eq:8} into a sequence of simpler problems through an ADMM algorithm~\cite{Boyd2011}. The augmented Lagrangian functional associated to~\eqref{eq:8} is defined as 
\begin{equation}
\begin{array}{l}
\mathcal{L}_\beta(z,\theta,\lambda) =    \displaystyle\sum_{c=1}^C \left\| \ba^c(\theta) \cdot \begin{bmatrix} z_x \\ z_y \end{bmatrix} - b^c(\theta) \right\|_2^2 \\
\qquad + \langle \lambda, (z_x,z_y) - \theta \rangle + \frac{\beta}{2} \left\| (z_x,z_y) - \theta \right\|_2^2,
\end{array}
\end{equation}
where $\lambda$ represent Lagrange multipliers, and $\beta>0$. ADMM then minimizes~\eqref{eq:8} by the following iterations:
\begin{align} 
\!z^{(k+1)}  & = \underset{z}{\operatorname{argmin}~} \mathcal{L}_{\beta^{(k)}}(z,\theta^{(k)},\lambda^{(k)}), \label{eq:10} \\
\theta^{(k+1)}  & = \underset{\theta}{\operatorname{argmin}~} \mathcal{L}_{\beta^{(k)}}(z^{(k+1)},\theta,\lambda^{(k)}), \label{eq:11} \\
\lambda^{(k+1)} & = \!\lambda^{(k)} \!\!+\! \beta^{(k)} \!\!\left(\! (z^{(k+1)}_x,\!z^{(k+1)}_y) \!-\! \theta^{(k+1)} \!\right). 
\end{align}
where $\beta^{(k)}$ is determined automatically~\cite{He2000}.

We then discretize~\eqref{eq:10} by finite differences, and solve the discrete optimality conditions by conjugate gradient. With this approach, no explicit boundary condition is required. As for~\eqref{eq:11}, it is solved in each pixel by a Newton method~\cite{Coleman1996}. In our experiments, the algorithm stops when the relative residual of the energy in~\eqref{eq:7} falls below $10^{-3}$.

\subsection{Experiments}

Since our method estimates a locally optimal solution, initialization matters. There is one situation where a reasonable initial estimate is available. This is when using an RGB-D camera: the depth channel D is noisy, but it may be refined using shading~\cite{Chatterjee2015,Choe2017,Han2013,Or-El2016,Or-El2015,Wu2014}. Hence, to qualitatively evaluate our approach, we consider in Figure~\ref{fig:RGBD} three real-world RGB-D datasets from~\cite{Han2013}, estimating lighting from the rough depth map (assuming $\rho \equiv 1$).  We attain the finest level of surface detail possible, since the surface is not over-smoothed through regularization.

For quantitative evaluation, we use in Figure~\ref{fig:experiments} the well-know ``Joyful Yell'' dataset, using three lighting scenarios. We first consider greylevel images, with a single-order and then a second-order lighting model, respectively defined by:
\begin{align}
  &\bl^1 =  [0.1,-0.25,-0.7,0.2,0,0,0,0,0]^\top, \label{eq:l1} \\
  &\bl^2 = [0.2,0.3,-0.7,0.5,-0.2,-0.2,0.3,0.3,0.2]^\top \label{eq:l2}. 
\end{align}
In the third experiment, we consider a colored, second-order lighting model defined by:
\begin{equation}
  \bl^3 \!=\! \!\begin{bmatrix}
    -0.2\!&\!-0.2\!&\!-1\!&\!0.4\!&\!0.1\!&\!-0.1\!&\!-0.1\!&\!-0.1\!&\!0.05 \\
    0\!&\!0.2\!&\!-1\!&\!0.3\!&\!0\!&\!0.2\!&\!0.1\!&\!0\!&\!0.1 \\
    0.2\!&\!-0.2\!&\!-1\!&\!0.2\!&\!-0.1\!&\!0\!&\!0\!&\!0.1\!&\!0
  \end{bmatrix}^{\!\!\top}\!\!\!.
  \label{eq:l3}
\end{equation}

\newpage

The importance of initialization is assessed by using two different initial estimates. The accuracy of 3D-reconstruction is evaluated by the mean angular error between the recovered normals and the ground truth ones, and the ability to explain the input image is measured through the RMSE between the data and the image simulated from the 3D-reconstruction. 

We compared those values against SIRFS~\cite{Barron2015}, which is the only method for SFS under natural illumination whose code is freely available. For fair comparison, we disabled albedo and lighting estimation in SIRFS, and gave a zero weight to all smoothing terms. To avoid the artifacts shown in Figure~\ref{fig:5}, SIRFS's multi-scale strategy was used. Figure~\ref{fig:experiments} proves that SFS under natural illumination can be solved using a purely data-driven strategy, without resorting neither to regularization nor to multi-scale. Besides, the runtimes of our method and SIRFS are comparable: a few minutes in all cases, for images with $150.000$ non-black pixels.

Still, these experiments show that the proposed method strongly depends on the choice of the initial estimate. We now show how to better constrain the 3D-reconstruction problem through sparse multi-view correspondences.

\section{Multi-view Shape-from-shading}
\label{sec:disambiguation}

Although colored natural illumination partly disambiguates SFS, it does not entirely remove ambiguities~\cite{Johnson2011}. Another disambiguation strategy must be considered in the absence of a good initial estimate. We now show that sparse correspondences in a multi-view framework can be employed for this purpose. 

To this end, let us now assume that we are given $N$ images $\{I_i:\, \Omega_i \subset \RR^2 \to \RR^C\}_{i \in \{1,\dots,N\}}$, along with the corresponding albedo maps and lighting vectors, both assumed to be channel- and image-dependent and denoted by $\{\rho_i:\,\Omega \to \RR^C\}_i$ and  $\{\bl^c_i\}_{c,i}$. The joint resolution of the $N$ SFS problems could be achieved by solving $N$ variational problems such as~\eqref{eq:8}. However, this would result in $N$ inconsistent depth maps: the $N$ SFS problems need to be coupled.

\subsection{Sparse Multi-view Constraints}

We use multi-view consistency to couple the $N$ SFS problems, and show that ambiguities are limited when introducing sparse correspondences between the images. We conjecture that any ambiguity even disappears if the correspondence set is dense. This conjecture could probably be proved by following~\cite{Chambolle1994}, but we leave this as future work. 

Let us assume that some sparse inter-images pixel correspondences are given (which can be obtained, for instance, by matching SIFT descriptors), and let us write them as the following $\Omega_i \times \Omega_j \to \RR$ functions, where $\Omega_i$ and $\Omega_j$ are the masks of the object in images $i$ and $j$, $i<j$: 
\begin{equation}
  c_{i,j}(\bp_i,\bp_j) = \begin{cases} 
    1 \text{~if pixel~}\bp_i\text{~in image~}I_i\text{~is matched} \\
    ~\,\text{~with pixel~}\bp_j\text{~in image~}I_j, \\ 
    0 \text{~otherwise}. 
  \end{cases}
\end{equation} 

Assuming perspective projection, a 3D-point $\bx$ in world coordinates is conjugate to a pixel $\bp_i$ according to
\begin{equation}
\bx = e^{z_i(\bp_i)} \bR_i \left[\frac{1}{f_i}{\tilde{\bp}_i}^\top,1\right]^\top + \bt_i,~\forall \bp_i \in \Omega_i,
\end{equation}
where $e^{z_i}$ is the $i$-th depth map (recall that we set $z_i$ to the $\log$ depth map under perspective projection), $\tilde{\bp}_i$ is the pixel coordinates \wrt the $i$-th principal point, $f_i$ is the $i$-th focal length, and $\bR_i \in \RR^{3 \times 3}$ and $\bt_i \in \RR^3$ are the rotation and translation describing the $i$-th pose of the camera (we assume that these poses are calibrated).

The multi-view consistency constraint then writes
\begin{align}
& c_{i,j}(\bp_i,\bp_j) \! \left( \! e^{z_i({\bp}_i)} \bR_i \!\left[\frac{1}{f_i} \tilde{\bp}_i^\top,1\right]^\top \!-\!  e^{z_j(\bp_j)} \bR_j \! \left[\frac{1}{f_j} \tilde{\bp}_j^\top,1\right]^\top \! \right)  \nonumber \\
& \quad -  c_{i,j}(\bp_i,\bp_j) \left( \bt_j - \bt_i \right) = {\bm 0},
\label{eq:15}
\end{align} 
which we rewrite as the following nonlinear constraint:
\begin{equation}
\bC_{i,j}(z_i,z_j) - \bd_{i,j} = 0,
\label{eq:16}
\end{equation}
where $\bC_{i,j}(z_i,z_j)$ is a $\Omega_i \times \Omega_j \to \RR^3$ function depending on the depth maps $z_i$ and $z_j$, whereas the function $\bd_{i,j}:\,\Omega_i \times \Omega_j \to \RR^3$  does not. 

\subsection{Proposed Variational Paradigm}

To disambiguate SFS through multi-views, we suggest to use $\mathcal{G}(z_i,z_j) = \lambda \left\| \bC_{i,j}(z_i,z_j) - \bd_{i,j} \right\|^2_{2,i,j}$ in the variational model~\eqref{eq:MV_SFS}, where $\|\cdot \|_{2,i,j}$ is the $\ell^2$ norm over $\Omega_i \times \Omega_j$, and $\lambda \geq 0$ is a weighting factor. Since the constraint~\eqref{eq:16} only depends on the depth values, and not on their gradients, we rather write it in terms of the auxiliary variables of the ADMM algorithm. This is motivated by the fact that the updates of these variables already require per-pixel nonlinear least-squares optimization. Moreover, the depth updates remain linear least-squares ones if the multi-view constraint is written in terms of the auxiliary variables. We thus define new auxiliary variables $\theta_i = ((z_i)_x,(z_i)_y,z_i)$, and turn~\eqref{eq:8} into: 
\begin{equation}
\begin{array}{l}
 \!\! \underset{ \substack{ \{z_i:\,\Omega_i \to \RR\}_i \\ \{\theta_i:\, \Omega_i \to\RR^3\}_i }  }{\min~}  \! \displaystyle\sum_{i=1}^N \sum_{c=1}^C \left\| \ba^c(\theta_i) \!\cdot\! \begin{bmatrix} {(z_i)}_x \\ {(z_i)}_y \end{bmatrix} \!-\! b^c(\theta_i) \right\|_{2,i}^2  \\
  \qquad \qquad + \dfrac{\lambda}{2} \displaystyle\mathop{\sum\sum}_{1 \leq i<j \leq N} \left\| \bC_{i,j}(\theta_i,\theta_j) - \bd_{i,j} \right\|_{2,i,j}^2 \\
  \text{s.t.}~ ({(z_i)}_x,{(z_i)}_y,z_i) = \theta_i,~\forall i \in \{1,\dots,N\}.
\end{array}
\label{eq:17}
\end{equation}

\begin{figure*}[!ht]
\centering
  \setlength{\tabcolsep}{0.1em}
  \begin{tabular}{ccccc}
    \includegraphics[height=0.25\linewidth]{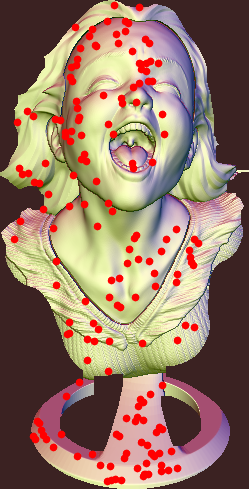} &
    \includegraphics[height=0.25\linewidth]{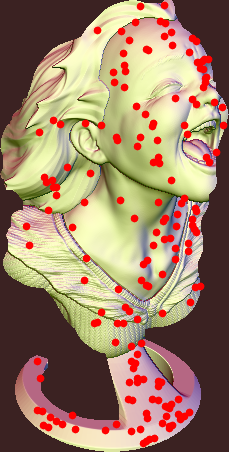} &
    \includegraphics[height=0.25\linewidth,trim = 8em 6em 8em 4em,clip]{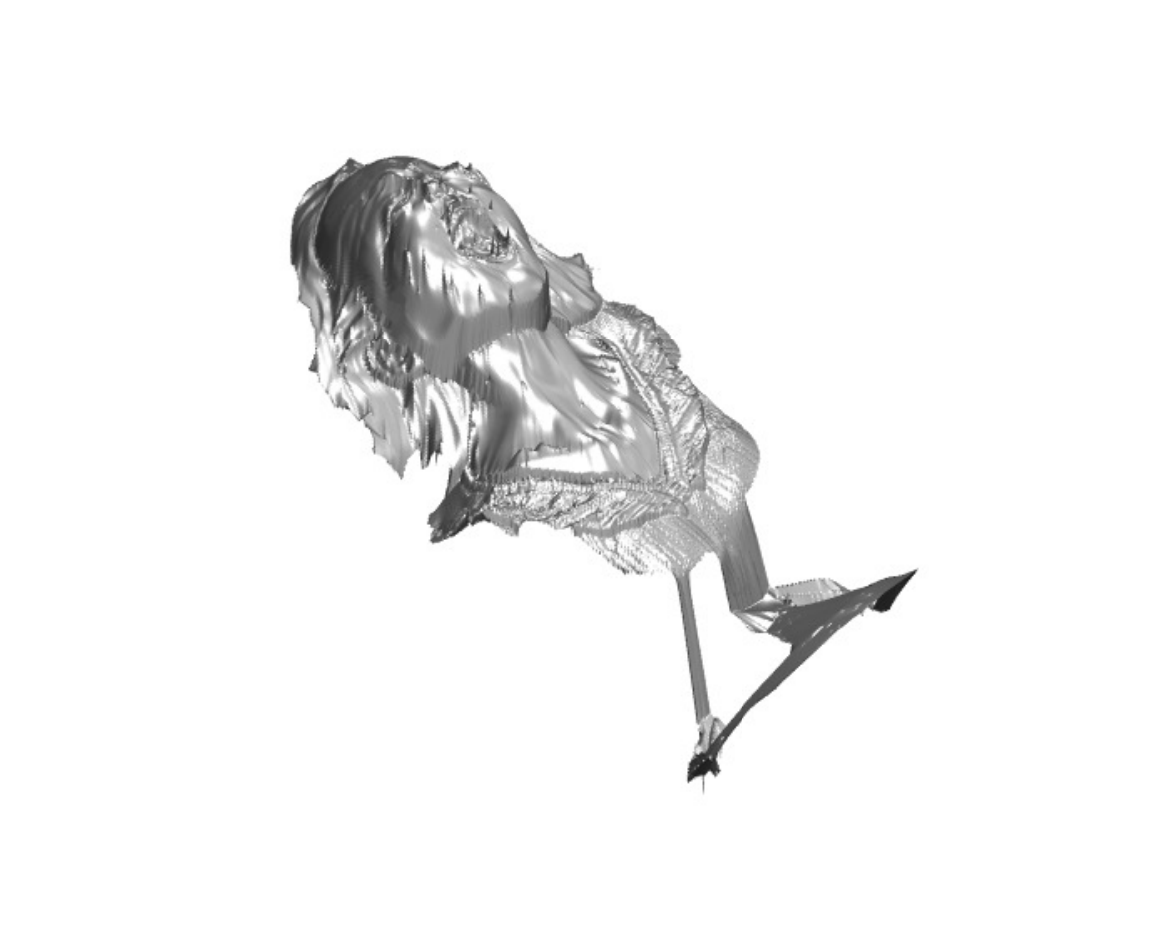} &
    \includegraphics[height=0.25\linewidth,trim = 8em 6em 8em 4em,clip]{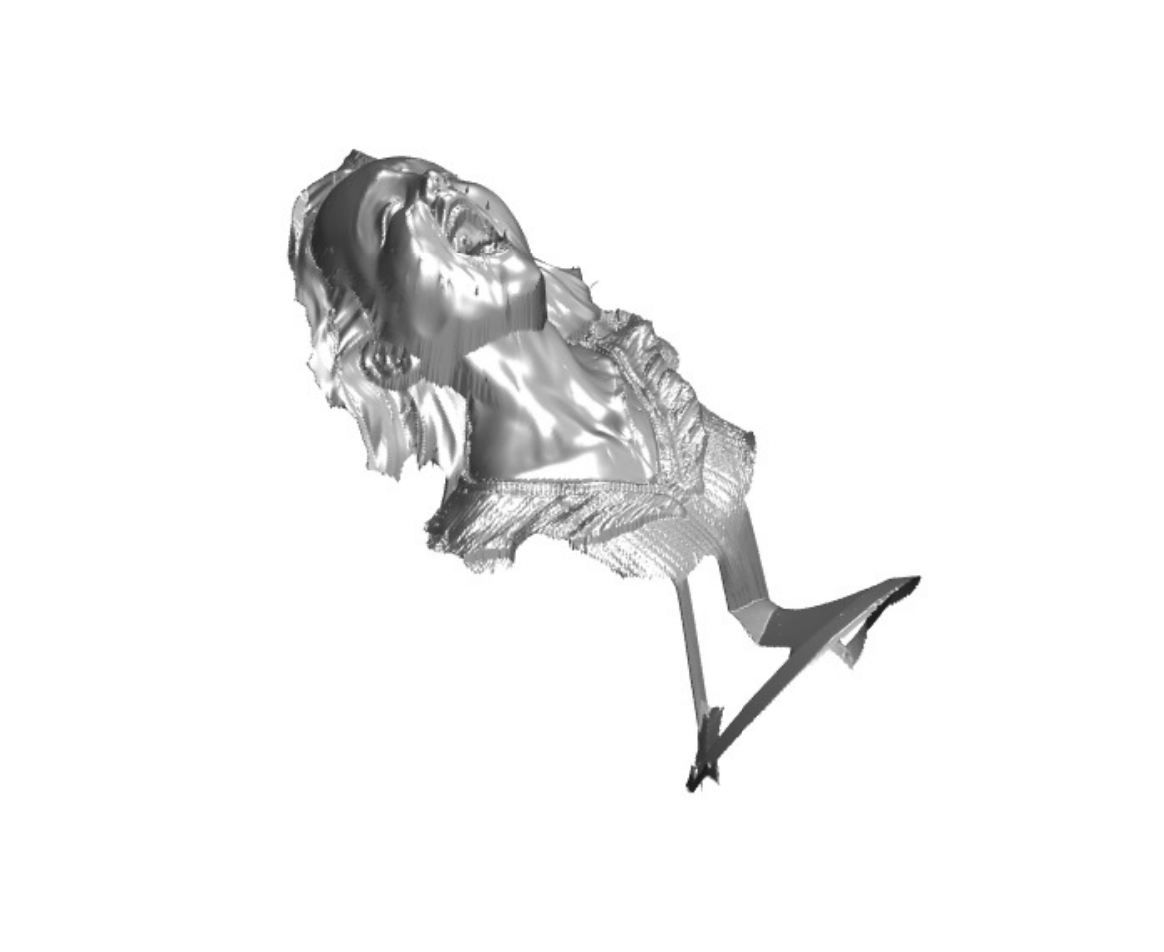} & 
    \includegraphics[height=0.25\linewidth]{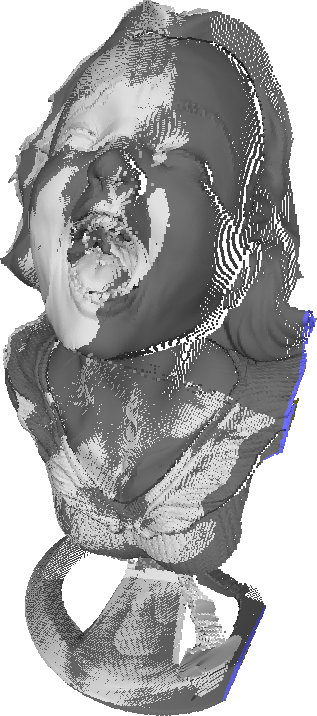} \\
     {\small $I_1$} & {\small $I_2$} &
      {\small Greylevel, first-order lighting} &
       {\small Colored, second-order lighting} & 
        {\small Fused point cloud} \\
   & & {\small  MAE-N $= 18.34$} & {\small MAE-N $= 8.13$}              
  \end{tabular}
  \caption{Binocular shape-from-shading. Left: input synthetic images and sparse correspondences, under the same colored, second-order lighting as in Figure~\ref{fig:experiments}. With $N=2$ views, SFS is disambiguated (no initial estimate is needed) and the 3D-reconstruction error is largely decreased. On the right, we show the point cloud obtained by fusing both depth maps $z_1$ and~$z_2$ obtained from the color 3D-reconstruction. }
  \label{fig:MV}
\end{figure*}

We experimentally found that the choice of a particular value of the parameter $\lambda$ is not important. Obviously, if $\lambda$ is set to $0$, then the $N$ SFS are uncoupled, and thus ambiguous. Yet, as long as $\lambda$ is ``high enough'', ambiguities disappear. In our tests, we found that the range $\lambda \in [10^{-8},10^{-2}]$ provides comparable results, and always used the value $\lambda = 10^{-5}$.

It is straightforward to modify the previous ADMM algorithm for solving~\eqref{eq:17}. In Figure~\ref{fig:MV}, we show the 3D-reconstructions obtained from $N=2$ synthetic views, in the same lighting scenarios as in the first and third experiments of Figure~\ref{fig:experiments}, using the same non-realistic initial estimate. We used $173$ pixel correspondences which were randomly picked using the ground-truth geometry. In comparison with the single-view results (see Figure~\ref{fig:experiments}), the estimated depth maps are more accurate. Besides, if we fuse both depth maps into a point cloud (using the known camera poses), we observe that both 3D-reconstructions are ``consistent'', which proves that amiguities are eliminated. 

\begin{figure}[!ht]
\centering
  \setlength{\tabcolsep}{0.1em}
  \begin{tabular}{cc}
   \includegraphics[width=0.5\linewidth]{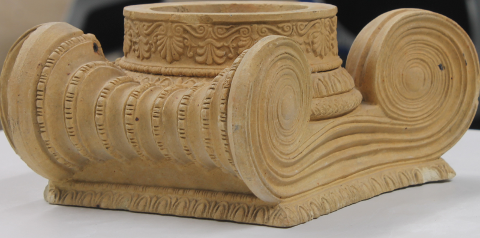}\hspace*{-2.35em}\includegraphics[height=0.1\linewidth]{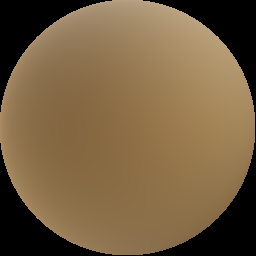} &
   \includegraphics[width=0.4\linewidth]{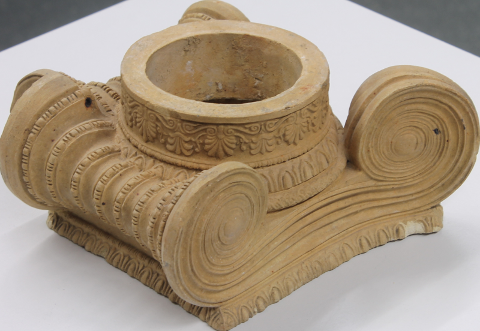}\hspace*{-2.35em}\includegraphics[height=0.1\linewidth]{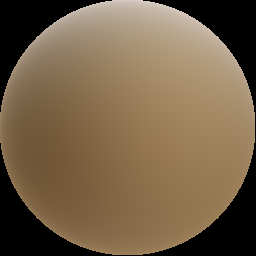} 
  \end{tabular}
  \begin{tabular}{cc}
   \includegraphics[width=0.48\linewidth]{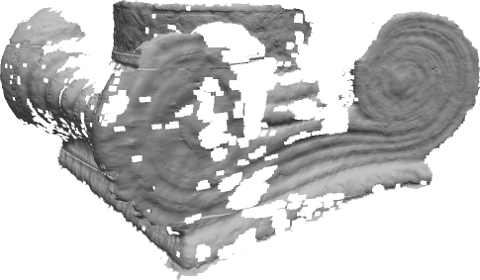} &
   \includegraphics[width=0.48\linewidth]{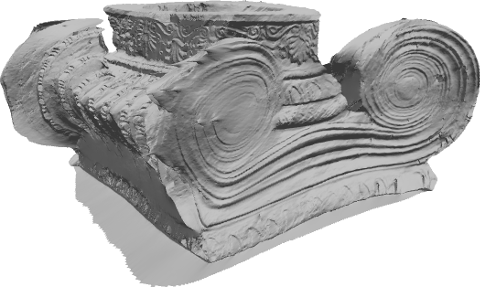} \\ 
   {\small CMPMVS~\cite{Jancosek2011} ($N=30$ views)} & {\small Ours ($N=2$ views)}
  \end{tabular}
\caption{3D-reconstruction of the ``Figure'' object~\cite{Zollhoefer2015}. Top: $N=2$ input real images $I_1$ and $I_2$. Bottom: depth map $z_1$ estimated using MVS (left, we show one out of $N=30$ depth maps from CMPMVS, before meshing~\cite{Jancosek2011}), and the proposed multi-view SFS method (right). The latter yields a much more dense reconstruction with more fine-scale details, although only $N=2$ views are used.}
\label{fig:figure}
\end{figure}

Eventually, we present in Figures~\ref{fig:teaser} and~\ref{fig:figure} the results of our method on two real-world datasets from~\cite{Zollhoefer2015}. We chose these datasets because they exhibit a uniform, though unknown, albedo. This albedo can thus be estimated during lighting calibration (since illumination is not provided in these datasets, it was calculated from the 3D-reconstructions provided in~\cite{Zollhoefer2015}, but these 3D-reconstructions were then not used any further). Sparse correspondences were extracted by matching standard SIFT features~\cite{Lowe1999} (the total number of used matches is worth $7982$ for the $N=4$ images of the ``Sokrates'' dataset, and $162$ for the $N=2$ images of the ``Figure'' one). These real-world experiments demonstrate that shading-based multi-view 3D-reconstruction constitutes a promising alternative to standard dense multi-view stereo.

\section{Conclusion}

We have shown how to achieve dense multi-view 3D-reconstruction without dense correspondences. A new variational approach to shape-from-shading under general lighting is used as the main tool for densification. It allows to drastically reduce the number of required images, while improving the amount of detail in the 3D-reconstruction. In future work, the new approach may be extended by automatic estimation of the albedo and of the lighting. This would allow coping with a broader variety of surfaces, and simplify the overall procedure. 

{\small
\bibliographystyle{ieee}
\bibliography{biblio}

\begin{thebibliography}{10}\itemsep=-1pt

\bibitem{Ackermann2015}
J.~Ackermann and M.~Goesele.
\newblock A survey of photometric stereo techniques.
\newblock {\em Foundations and Trends in Computer Graphics and Vision},
  9(3-4):149--254, 2015.

\bibitem{Barron2015}
J.~T. Barron and J.~Malik.
\newblock Shape, illumination, and reflectance from shading.
\newblock {\em IEEE Transactions on Pattern Analysis and Machine Intelligence},
  37(8):1670--1687, 2015.

\bibitem{Basri2003}
R.~Basri and D.~P. Jacobs.
\newblock {Lambertian reflectances and linear subspaces}.
\newblock {\em IEEE Transactions on Pattern Analysis and Machine Intelligence},
  25(2):218--233, 2003.

\bibitem{Blake1985}
A.~Blake, A.~Zisserman, and G.~Knowles.
\newblock Surface descriptions from stereo and shading.
\newblock {\em Image and Vision Computing}, 3(4):183--191, 1985.

\bibitem{Boyd2011}
S.~Boyd, N.~Parikh, E.~Chu, B.~Peleato, and J.~Eckstein.
\newblock {Distributed Optimization and Statistical Learning via the
  Alternating Direction Method of Multipliers}.
\newblock {\em Foundations and Trends in Machine Learning}, 3(1):1--122, 2011.

\bibitem{Breuss2012}
M.~Breu{\ss}, E.~Cristiani, J.-D. Durou, M.~Falcone, and O.~Vogel.
\newblock Perspective shape from shading: Ambiguity analysis and numerical
  approximations.
\newblock {\em SIAM Journal on Imaging Sciences}, 5(1):311--342, 2012.

\bibitem{Bruss1982}
A.~R. Bruss.
\newblock The eikonal equation: Some results applicable to computer vision.
\newblock {\em Journal of Mathematical Physics}, 23(5):890--896, 1982.

\bibitem{Chambolle1994}
A.~Chambolle.
\newblock A uniqueness result in the theory of stereo vision: coupling shape
  from shading and binocular information allows unambiguous depth
  reconstruction.
\newblock {\em Annales de l'IHP - Analyse non lin{\'e}aire}, 11(1):1--16, 1994.

\bibitem{Chatterjee2015}
A.~Chatterjee and V.~Madhav~Govindu.
\newblock Photometric refinement of depth maps for multi-albedo objects.
\newblock In {\em Proceedings of CVPR}, 2015.

\bibitem{Choe2017}
G.~Choe, J.~Park, Y.-W. Tai, and I.~S. Kweon.
\newblock Refining geometry from depth sensors using {IR} shading images.
\newblock {\em International Journal of Computer Vision}, 122(1):1--16, 2017.

\bibitem{Coleman1996}
T.~F. Coleman and Y.~Li.
\newblock An interior trust region approach for nonlinear minimization subject
  to bounds.
\newblock {\em SIAM Journal on Optimization}, 6(2):418--445, 1996.

\bibitem{Cristiani2007}
E.~Cristiani and M.~Falcone.
\newblock Fast semi-lagrangian schemes for the eikonal equation and
  applications.
\newblock {\em SIAM Journal on Numerical Analysis}, 45(5):1979--2011, 2007.

\bibitem{Durou2008}
J.-D. Durou, M.~Falcone, and M.~Sagona.
\newblock Numerical methods for shape-from-shading: A new survey with
  benchmarks.
\newblock {\em Computer Vision and Image Understanding}, 109(1):22--43, 2008.

\bibitem{Frolova2004}
D.~Frolova, D.~Simakov, and R.~Basri.
\newblock Accuracy of spherical harmonic approximations for images of
  lambertian objects under far and near lighting.
\newblock In {\em Proceedings of ECCV}, 2004.

\bibitem{Furukawa2015}
Y.~Furukawa and C.~Hern{\'a}ndez.
\newblock Multi-view stereo: A tutorial.
\newblock {\em Foundations and Trends in Computer Graphics and Vision},
  9(1-2):1--148, 2015.

\bibitem{Galliani2016}
S.~Galliani and K.~Schindler.
\newblock Just look at the image: Viewpoint-specific surface normal prediction
  for improved multi-view reconstruction.
\newblock In {\em Proceedings of CVPR}, 2016.

\bibitem{Han2013}
Y.~Han, J.-Y. Lee, and I.~S. Kweon.
\newblock High quality shape from a single {RGB-D} image under uncalibrated
  natural illumination.
\newblock In {\em Proceedings of ICCV}, 2013.

\bibitem{He2000}
B.~S. He, H.~Yang, and S.~L. Wang.
\newblock Alternating direction method with self-adaptive penalty parameters
  for monotone variational inequalities.
\newblock {\em Journal of Optimization Theory and Applications},
  106(2):337--356, 2000.

\bibitem{Hernandez2007}
C.~Hern{\'a}ndez, G.~Vogiatzis, G.~J. Brostow, B.~Stenger, and R.~Cipolla.
\newblock {Non-rigid Photometric Stereo with Colored Lights}.
\newblock In {\em {Proceedings of ICCV}}, 2007.

\bibitem{Horn1986}
B.~K.~P. Horn and M.~J. Brooks.
\newblock The variational approach to shape from shading.
\newblock {\em Computer Vision, Graphics, and Image Processing},
  33(2):174--208, 1986.

\bibitem{Horn1989a}
B.~K.~P. Horn and M.~J. Brooks, editors.
\newblock {\em {Shape from Shading}}.
\newblock MIT Press, 1989.

\bibitem{Huang2011}
R.~Huang and W.~A.~P. Smith.
\newblock Shape-from-shading under complex natural illumination.
\newblock In {\em Proceedings of ICIP}, 2011.

\bibitem{Ikeuchi1981}
K.~Ikeuchi and B.~K. Horn.
\newblock Numerical shape from shading and occluding boundaries.
\newblock {\em Artificial intelligence}, 17(1-3):141--184, 1981.

\bibitem{Jancosek2011}
M.~Jancosek and T.~Pajdla.
\newblock Multi-view reconstruction preserving weakly-supported surfaces.
\newblock In {\em Proceedings of CVPR}, 2011.

\bibitem{Jin2008}
H.~Jin, D.~Cremers, D.~Wang, A.~Yezzi, E.~Prados, and S.~Soatto.
\newblock {3-D Reconstruction of Shaded Objects from Multiple Images Under
  Unknown Illumination}.
\newblock {\em International Journal of Computer Vision}, 76(3):245--256, 2008.

\bibitem{Johnson2011}
M.~K. Johnson and E.~H. Adelson.
\newblock Shape estimation in natural illumination.
\newblock In {\em Proceedings of CVPR}, 2011.

\bibitem{Kim2016}
K.~Kim, A.~Torii, and M.~Okutomi.
\newblock {Multi-view Inverse Rendering Under Arbitrary Illumination and
  Albedo}.
\newblock In {\em Proceedings of ECCV}, 2016.

\bibitem{Kolev2009}
K.~Kolev, M.~Klodt, T.~Brox, and D.~Cremers.
\newblock Continuous global optimization in multview 3d reconstruction.
\newblock {\em International Journal of Computer Vision}, 2009.

\bibitem{Langguth2016}
F.~Langguth, K.~Sunkavalli, S.~Hadap, and M.~Goesele.
\newblock {Shading-aware Multi-view Stereo}.
\newblock In {\em Proceedings of ECCV}, 2016.

\bibitem{Lions1993}
P.-L. Lions, E.~Rouy, and A.~Tourin.
\newblock Shape-from-shading, viscosity solutions and edges.
\newblock {\em Numerische Mathematik}, 64(1):323--353, 1993.

\bibitem{Lowe1999}
D.~G. Lowe.
\newblock Object recognition from local scale-invariant features.
\newblock In {\em Proceedings of ICCV}, 1999.

\bibitem{Maurer2016}
D.~Maurer, Y.~C. Ju, M.~Breu{\ss}, and A.~Bruhn.
\newblock {Combining Shape from Shading and Stereo: A Variational Approach for
  the Joint Estimation of Depth, Illumination and Albedo}.
\newblock In {\em Proceedings of BMVC}, 2016.

\bibitem{Nehab2005}
D.~Nehab, S.~Rusinkiewicz, J.~Davis, and R.~Ramamoorthi.
\newblock Efficiently combining positions and normals for precise {3D}
  geometry.
\newblock {\em ACM Transactions on Graphics}, 24(3):536--543, 2005.

\bibitem{Or-El2016}
R.~Or-El, R.~Hershkovitz, A.~Wetzler, G.~Rosman, A.~M. Bruckstein, and
  R.~Kimmel.
\newblock Real-time depth refinement for specular objects.
\newblock In {\em Proceedings of CVPR}, 2016.

\bibitem{Or-El2015}
R.~Or-El, G.~Rosman, A.~Wetzler, R.~Kimmel, and A.~Bruckstein.
\newblock {RGBD-Fusion: Real-Time High Precision Depth Recovery}.
\newblock In {\em Proceedings of CVPR}, 2015.

\bibitem{Prados2003}
E.~Prados and O.~Faugeras.
\newblock Perspective shape from shading and viscosity solutions.
\newblock In {\em Proceedings of ICCV}, 2003.

\bibitem{Prados2005}
E.~Prados and O.~Faugeras.
\newblock Shape from shading: A well-posed problem?
\newblock In {\em Proceedings of CVPR}, 2005.

\bibitem{Ramamoorthi2001}
R.~Ramamoorthi and P.~Hanrahan.
\newblock {An Efficient Representation for Irradiance Environment Maps}.
\newblock In {\em Proceedings of SIGGRAPH}, 2001.

\bibitem{Richter2015}
S.~R. Richter and S.~Roth.
\newblock Discriminative shape from shading in uncalibrated illumination.
\newblock In {\em Proceedings of CVPR}, 2015.

\bibitem{Rouy1992}
E.~Rouy and A.~Tourin.
\newblock A viscosity solutions approach to shape-from-shading.
\newblock {\em SIAM Journal on Numerical Analysis}, 29(3):867--884, 1992.

\bibitem{Samaras2000}
D.~Samaras, D.~Metaxas, P.~Fua, and Y.~G. Leclerc.
\newblock Variable albedo surface reconstruction from stereo and shape from
  shading.
\newblock In {\em Proceedings of CVPR}, 2000.

\bibitem{Tankus2003}
A.~Tankus, N.~A. Sochen, and Y.~Yeshurun.
\newblock A new perspective (on) shape-from-shading.
\newblock In {\em Proceedings of ICCV}, 2003.

\bibitem{Tola2010}
E.~Tola, V.~Lepetit, and P.~Fua.
\newblock {DAISY: An Efficient Dense Descriptor Applied to Wide Baseline
  Stereo}.
\newblock {\em IEEE Transactions on Pattern Analysis and Machine Intelligence},
  32(5):815--830, 2010.

\bibitem{Vogiatzis2005}
G.~Vogiatzis, P.~Torr, and R.~Cippola.
\newblock Multi-view stereo via volumetric graph-cuts.
\newblock In {\em Proceedings of CVPR}, 2005.

\bibitem{Wu2011}
C.~Wu, B.~Wilburn, Y.~Matsushita, and C.~Theobalt.
\newblock High-quality shape from multi-view stereo and shading under general
  illumination.
\newblock In {\em Proceedings of CVPR}, 2011.

\bibitem{Wu2014}
C.~Wu, M.~Zollh\"{o}fer, M.~Nie{\ss}ner, M.~Stamminger, S.~Izadi, and
  C.~Theobalt.
\newblock Real-time shading-based refinement for consumer depth cameras.
\newblock {\em ACM Transactions on Graphics}, 33(6):200:1--200:10, 2014.

\bibitem{Zhang1999}
R.~Zhang, P.-S. Tsai, J.~E. Cryer, and M.~Shah.
\newblock Shape-from-shading: a survey.
\newblock {\em IEEE Transactions on Pattern Analysis and Machine Intelligence},
  21(8):690--706, 1999.

\bibitem{Zollhoefer2015}
M.~Zollh{\"o}fer, A.~Dai, M.~Innman, C.~Wu, M.~Stamminger, C.~Theobalt, and
  M.~Nie{\ss}ner.
\newblock Shading-based refinement on volumetric signed distance functions.
\newblock {\em ACM Transactions on Graphics}, 34(4):96:1--96:14, 2015.

\end{thebibliography}
}

\end{document}